\documentclass{article}

     \usepackage[preprint]{neurips_2025}

\usepackage{amsmath,amsfonts,bm}

\def\eqref#1{equation~\ref{#1}}

\def\1{\bm{1}}

\DeclareMathAlphabet{\mathsfit}{\encodingdefault}{\sfdefault}{m}{sl}
\SetMathAlphabet{\mathsfit}{bold}{\encodingdefault}{\sfdefault}{bx}{n}

\newcommand{\R}{\mathbb{R}}

\usepackage[utf8]{inputenc} 
\usepackage[T1]{fontenc}    
\usepackage{hyperref}       
\usepackage{url}            
\usepackage{booktabs}       
\usepackage{amsfonts}       
\usepackage{graphicx}       
\usepackage{amsthm}         
\usepackage{nicefrac}       
\usepackage{microtype}      
\usepackage{xcolor}         
\usepackage{thmtools,thm-restate}
\usepackage{amssymb} 
\usepackage{enumerate}
\usepackage{natbib}
\usepackage{comment}

\usepackage{bbm}

\newtheorem{theorem}{Theorem}

\newtheorem{lemma}[theorem]{Lemma}

\usepackage{algpseudocode}
\newcommand{\naturals}{\mathbb{N}}

\newcommand{\regret}{{\cal R}}

\newcommand{\mymid}{\,|\,}
\newcommand{\mydoublemid}{\,|| \,}

\title{Information-Theoretic Framework for Understanding Modern Machine-Learning}
\usepackage{times}

\author{Meir Feder
	\\
    School of Electrical and Computer Engineering \\
    Tel-Aviv University \\
    Tel-Aviv, Israel \\
	\texttt{meir@tau.ac.il} \\
    \And
	Ruediger Urbanke \\
    School of Computer and Communication Sciences \\
École Polytechnique Fédérale de Lausanne (EPFL) \\
Lausanne, Switzerland \\
    \texttt{rudiger.urbanke@epfl.ch} \\
    \And
    Yaniv Fogel \\
    School of Electrical and Computer Engineering \\
    Tel-Aviv University \\
    Tel-Aviv, Israel \\
	\texttt{yaniv.fogel8@gmail.com} \\
}
\begin{document}

\maketitle

\begin{abstract}
We introduce an {\em information-theoretic} framework that views learning as {\em universal prediction} under {\em log loss}, characterized through {\em regret bounds}. Central to the framework is an effective notion of {\em architecture-based model complexity}, defined by the probability mass or volume of models in the vicinity of the data-generating process, or its projection on the model class. This volume is related to spectral properties of the expected Hessian or the {\em Fisher Information Matrix}, leading to tractable approximations. We argue that successful architectures possess a \emph{broad complexity range}, enabling learning in highly over-parameterized model classes. The framework sheds light on the role of {\em inductive biases}, the effectiveness of {\em stochastic gradient descent}, and phenomena such as {\em flat minima}. It unifies {\em online}, {\em batch}, {\em supervised}, and {\em generative} settings, and applies across the {\em stochastic-realizable} and {\em agnostic} regimes. Moreover, it provides insights into the success of modern machine-learning architectures, such as deep neural networks and transformers, suggesting that their broad complexity range naturally arises from their {\em layered structure}. These insights open the door to the design of alternative architectures with potentially comparable or even superior performance. 
\end{abstract}


\section{Introduction}
We present a structured, {\em information-theoretic} framework for understanding learning in modern machine-learning architectures. Our perspective treats learning as a problem of \emph{universal prediction} under \emph{log loss}. An architecture or model class is represented as a set of {\em parameterized probability distributions}. For modern architectures such as deep neural networks (DNNs) or transformers, these distributions are typically characterized by a large number of parameters. 
Learning corresponds to constructing a universal probability distribution that serves as a surrogate for the unknown model (or the best unknown model) within the class. 

For large model classes, uniform regret bounds are often vacuous; therefore, we introduce a {\em nonuniform} notion of \emph{model complexity}, leading to nonuniform regret bounds. Our universal learner is Bayesian, optimal in a way formally described in Appendix \ref{sec:mixture_modelsdetails}, represented as a uniform (or almost uniform) \emph{mixture} over the models in the class, where the effective nonuniform complexity is governed by the total probability mass of models that are ``close'', in an information distance, to the best model in the class, which in the realizable case is the true data-generating process.

Because this complexity can be difficult to compute, we evaluate it locally and relate it to the spectrum of the {\em Hessian} or the {\em Fisher Information Matrix} (FIM), yielding a somewhat weaker but more tractable auxiliary complexity measure. We argue that for contemporary architectures, this complexity {\em aligns naturally with the statistical structure of real-world signals}, and that standard training procedures, most notably stochastic gradient descent (SGD), essentially implement the framework both effectively and efficiently. 

The formulation clarifies the role of \emph{architectural inductive biases}. Such biases are inherent in architectures such as deep neural networks and transformers. The formulation explains the source of these biases and derives, from first principles, mechanisms underlying empirical phenomena such as \emph{flat minima}.
It also suggests that other architectures that have similar composite structures can also have this {\em broad range of complexity} and can be valid alternatives to the current architectures.

To demonstrate the explanatory power of the framework, we present an illustrative numerical experiment that computes the spectral behavior of the relevant information measures. 
We also highlight large-scale studies in the literature whose findings agree with our predictions.

The framework applies broadly across settings common in modern ML. It encompasses both the \emph{online} and \emph{batch} regimes under the umbrella of \emph{prediction}, where the objective is to forecast the next symbol given the observed sequence. This paradigm lies at the core of generative AI and, in particular, of \emph{language models}. It also covers the \emph{supervised} setting, including standard classification and regression tasks.
While the \emph{realizable case} where data is generated by a model in the class is the most straightforward, the framework naturally extends to the more general \emph{agnostic case} where the true distribution may lie outside the class and even to the individual sequence setting, which makes no probabilistic assumptions at all.


In summary, this paper offers a unified, information-theoretic perspective that sheds light on why modern models with extremely large parameter count, often surpassing the size of their training datasets, can still succeed. The results are situated within a broad landscape of related research and supported by extensive empirical evidence. Section~\ref{sec:contributions_and_prior_work} outlines our main contributions and situates them in the context of prior work.

\section{Setup, Learning as Probabilistic Prediction, and Regret}
For simplicity, in the discussion below we consider {\em supervised learning}. All concepts carry over to the online and batch setting and we discuss those settings in detail in the corresponding appendices. To simplify further, we often omit labels in the main text to indicate that this relates to the supervised case.

In the standard supervised learning setting, the {\em model class} is defined as a collection of probability distributions that assign probabilities to possible outcomes $y$ given inputs $x$:
\begin{align}
\Theta = \Bigl\{ P_{\theta}(y \mymid x) : x \in {\mathcal X},\; y \in {\mathcal Y},\; P_{\theta}(y \mymid x) \geq 0,\; \sum_{y} P_{\theta}(y \mymid x) = 1 \Bigr\}_{\theta \in \Theta}. 
\label{equ:model_class_supervised}
\end{align}
Note that we write $\sum_{y} P_{\theta}(y \mymid x) = 1$ but our framework equally applies to continuous as well as to discrete distributions.
This generalizes the classical machine learning setting, where the model class consists of deterministic functions $y=f_\theta(x)$.

From an information-theoretic perspective, given the model class and a sample training set $\mathcal{S}=\{(x_i, y_i)\}_{i=1}^{n}$, \emph{learning} means constructing a predictive distribution $Q(y \mymid x; \mathcal{S})$ that assigns a probability distribution to all possible $y$ given a new input $x$.

We use the \emph{log-loss}, so 
for a test pair $(x, y)$, the learner incurs a loss of $-\log\bigl(Q(y \mymid x; \mathcal{S})\bigr)$.
The log-loss is the natural choice when the model class consists of probability distributions, because it is {\em strictly proper} \cite{GneitingRaftery2007}, it aligns with maximum likelihood estimation \cite{casella2002statistical}, it corresponds to the self-information or ``codelength'' of the resulting outcome \cite{MerhavFeder1998}, it corresponds to minimizing the Kullback--Leibler (KL) divergence between the true and predicted distributions \cite{CoverThomas2012}, and it ensures calibrated probabilities in practice \cite{niculescu2005predicting}. Note that the KL-divergence $D_{\text{KL}} (P(x)\mydoublemid Q(x)) = \mathbb{E}_{x \sim P(\cdot)}\log \frac{P(x)}{Q(x)}$ can be regarded as the average extra log-loss between $Q$ and $P$ assuming that $x\sim P$. Similarly, we can define $D_{\text{KL}}(P(y|x)\mydoublemid Q(y|x))$ which depends on $x$ and the conditional KL-divergence $D_{\text{KL}}(P(y|x)\mydoublemid Q(y|x)\mymid X)$ as its expectation over $X$.

Let $P(y \mymid x)$ denote the {\em true data-generating distribution}. Assume that the input $x$ is generated according to $P_X(x)$. It is important to emphasize that, although we assume the inputs $x$ follow some distribution $P_X(x)$, {\em the proposed learner does not depend on this distribution} but only on the observed training set $\mathcal{S}$. 
Define
\[
\theta_0 = \arg \min_{\theta \in \Theta} D_{\text{KL}}(P(y \mymid x) \mydoublemid P_{\theta}(y \mymid x) \mymid X) =  \arg \min_{\theta \in \Theta} \mathbb{E}_{x \sim P_X} \left[ D_{\text{KL}} \left( P(y \mymid x) \mydoublemid P_{\theta}(y \mymid x) \right) \right].
\]
That is, we are in the {\em agnostic} setting, where the unknown true data-generating distribution may not be in the model class and where $P_{\theta_0}$ is its best approximation, in KL-divergence sense, within the model class. For a given input $x$, the \emph{pointwise regret} of using $Q(y \mymid x; \mathcal{S})$ instead of $P_{\theta_0}(y \mymid x)$ is:
\[
-\log\bigl(Q(y \mymid x; \mathcal{S})\bigr)\;-\;\bigl(-\log\bigl(P_{\theta_0}(y \mymid x)\bigr)\bigr)
\;=\;\log\frac{P_{\theta_0}(y \mymid x)}{Q(y \mymid x; \mathcal{S})}.
\]
The \emph{expected regret}, averaged over both training and test samples, is
\begin{align}
\regret^s_{a}(Q,P) 
& = \mathbbm{E}_{\mathcal{S} \sim P_X P }\!\Bigl[\mathbbm{E}_{(x, y) \sim P_X P }\!\bigl[\log\bigl(\tfrac{P_{\theta_0}(y \mymid x)}{Q(y \mymid x; \mathcal{S})}\bigr)\bigr]\Bigr].
\label{equ:regret_supervised_agnostic}
\end{align}
The superscript $s$ stands for {\em supervised}, while the subscript $a$ in the notation above indicates that this is the {\em agnostic} case.
In the {\em realizable} case, where the true distribution lies in the model class we denote the data-generating distribution by $P_{\theta_0}$. The expression for the regret then simplifies to a conditional KL-divergence:
\begin{align}
\regret^s_{r}(Q,\theta_0) 
& = {D_{\text{KL}}}\bigl(P_{\theta_0}(Y \mymid X) \mydoublemid Q(Y \mymid X; {\mathcal S}) \mymid X; {\mathcal S}\bigr).
\label{equ:regret_supervised_realizable}
\end{align}

\section{Mixture Models}
\label{sec:mixture_models}
A principled {\em universal} learner in this setting is the \emph{Bayesian mixture distribution}:
\begin{align}
Q(y \mymid x; {\mathcal S}) = \int_{\theta} w(\theta \mid \mathcal{S}) P_{\theta}(y \mymid x) \, d\theta.
\label{equ:mixture_model_supervised}
\end{align}
Here, $w(\theta)$ is the prior distribution over the parameters, and the posterior distribution is computed according to $w(\theta \mid \mathcal{S}) \propto w(\theta) \, \prod_{i=1}^{n} P_{\theta}(y_i \mymid x_i)$.
It reweighs each hypothesis by how well it explains the observed data. We have made the standard assumption of iid data.
This approach produces a \emph{universal probability assignment} in the sense that it works without knowing which $P_{\theta}$ is closest to the true data-generating distribution, and it adapts automatically as more data are observed. 
Appendix~\ref{sec:mixture_modelsdetails} further discusses why mixture models make good learners.

\section{Regret Bounds}
\label{sec:regret_bounds}
For $\epsilon^2 > 0$, define the following region in the parameters space:
\begin{align*}
\Theta_0^{\epsilon^2} & = \left\{ \theta \in \Theta: \underbrace{\mathbbm{E}_{(x, y) \sim P_X(\cdot) P(\cdot \mymid \cdot)}\left[ \log \left( \frac{P_{\theta_0}(y \mymid  x)}{P_{\theta}(y \mymid x) }\right)\right]}_{=D_{\text{KL}}(P_{\theta_0}(y \mymid x) P_X(x) \mydoublemid P_{\theta}(y \mymid x) P_X(x)) \;\text{for realizable case}} \leq \epsilon^2 \right\}.
\end{align*}
Let $w(\Theta_0^{\epsilon^2} \mid \mathcal{S})=\int_{\Theta_0^{\epsilon^2}} w(\theta \mid \mathcal{S}) d\theta$ denote the posterior probability assigned to this region.  Then we can derive a {\em non-uniform} regret bound of the form:
\begin{align}
\regret^{s}_{a}(Q,P) \leq \epsilon^2 -  \mathbbm{E}_{{\mathcal S}}\left[\log w(\Theta_0^{\epsilon^2} \mymid \mathcal{S}) \right],
\label{equ:regretboundsupervised}
\end{align}
where ${\mathcal S}=\{(x_i, y_i)\}_{i=1}^{n} \sim \prod_{i=1}^{n} P_X(x_i) P(y_i \mymid x_i) $.
This bound is valid for \emph{any} $\epsilon^2 \geq 0$; choosing $\epsilon$ optimally yields the tightest bound. The bound (\ref{equ:regretboundsupervised}) is proved in Appendix \ref{sec:regret_bound_supervised}, while similar bounds for online and batch prediction are derived in the rest of Appendix \ref{sec:regret_boundsdetails}.


Our regret bound is {\em non-uniform} allowing it to break the following otherwise unavoidable packing-based limit $\text{(number of models)} \times \text{(probability per model)} \leq 1$.
E.g., in the classical information-theoretic approach, the quest for uniform min-max bounds leads to Jeffreys' prior over the parameter space, making the effective number of models in the class approximately equal to $ 2^C$, where $C$ is the Shannon capacity of the channel specified by the model class, resulting in $C$ as a uniform regret bound. 
This forces a trade-off between class size and tightness of bounds, which become useless for large $C$.  
Non-uniform bounds, in particular when their complexity range is broad, allow many models to have small probability while a subset has large probability, breaking this limitation.

\section{Model Complexity}
\label{sec:model_complexity}
The non-uniform regret bound motivates a non-uniform notion of model complexity:
\[
\mathrm{Comp}(P, {\epsilon^2}) = -  \mathbbm{E}_{{\mathcal S}}\left[\log w(\Theta_0^{\epsilon^2} \mid \mathcal{S}) \right].
\]
This complexity measure is:
\begin{itemize}
\setlength{\itemsep}{-0.1cm}
    \item \emph{Model-specific:} it does not only depend on the model class itself, but also on $P$, the data-generating distribution, and $P_{\theta_0}$, its projection on the model class.
    \item \emph{Data-dependent:} it depends on the structure of $\mathcal{S}$ and in particular on $|\mathcal{S}|$. It decreases as more data arrive, see e.g., Theorem \ref{the:supervised_local_properties_theorem}.
    \item \emph{Closely tied to regret:} simpler models yield tighter regret bounds - see Appendix \ref{sec:regret_boundsdetails}.
\end{itemize}

\paragraph{Prior complexity and complexity range}
When $\mathcal{S}=\varnothing$, no training,
the complexity reduces to:
\begin{align}
\mathrm{Comp}_{\mathrm{prior}}(P, \epsilon^2)  = - \log w(\Theta_0^{\epsilon^2}), \label{equ:pior_complexity}
\end{align}
the \emph{prior complexity}, i.e., the negative log prior probability of being within an $\epsilon^2$ neighborhood of the best model in the model class, which is $P_{\theta_0}$, the projection of $P$ on the model class. 
This quantity reflects the architecture’s \emph{implicit bias}: models with high prior probability in their neighborhood are favored and those models are easier to learn.

When the distribution of prior complexities within a model class is wide, the class is said to have a \emph{large complexity range}. Such a class can be both highly expressive, capable of representing many functions, while still assigning sufficient preference to simpler models to achieve tight regret bounds.
Deep neural networks and transformers generally exhibit a large complexity range, whereas linear models have virtually none. In Section~\ref{sec:theroleofthearchitecture}, we examine the characteristics of “good” architectures that possess a broad complexity range.

\paragraph{Evidence accumulation.}
For $n \geq 1$ data points, complexity roughly decomposes into prior complexity plus an \emph{evidence term} from the likelihood.  
Under small $\epsilon$, the integral over $\Theta_0^{\epsilon^2}$ approximately factorizes into prior $\times$ likelihood, so:
\[
\mathrm{Comp}(P, {\epsilon^2}) \approx \mathrm{Comp}_{\mathrm{prior}}(P, \epsilon^2) + \text{(evidence term)}.
\]
Note that the regret is upper bounded by $2 \max \{\epsilon^2, \mathrm{Comp}(P, {\epsilon^2})\}$.

Computing the exact complexity can be cumbersome or even infeasible for large models.
Two useful proxies are:
\begin{enumerate}
\setlength{\itemsep}{-0.1cm}
\item \textit{The minimal number of neurons (or parameters) required to represent $P_{\theta_0}$.}
Concrete examples of this relationship are given in \cite{GluchUrbanke2023}. Although the related computations may be intricate, the underlying principle is straightforward. This measure is closely related to prior complexity: if a function can be represented with only a few active neurons, many parameters can vary freely (thanks to non-linearities) without altering the function. Consequently, the prior assigns high probability mass to such representations.  In Appendix~\ref{app:zero_function_has_small_complexity} we discuss a simple example which illustrates this point. We consider the ``zero'' function in a ReLU network of depth one and derive a concrete lower bound on its prior probability, showing that indeed it has low prior complexity.

\item \textit{The spectrum of the expected Hessian or the Fisher Information Matrix (FIM).}
This perspective is closely related to the idea of \emph{flat minima}, which empirical studies (e.g., \cite{HochreiterSchmidhuber1997}) have associated with strong generalization. Our framework refines this view: it is not flatness per se that ensures generalization, but the fact that flat regions correspond to large volumes of $\Theta_0$. A larger volume implies lower prior complexity, which in turn yields lower regret.

More concretely, consider the spectrum of the expected Hessian of the log-score or the FIM. Large eigenvalues identify directions in the parameter space that are essential for representing the target function or distribution, and thus determine the number of effective parameters. In contrast, small eigenvalues indicate directions along which parameters can vary freely without significantly altering the function or distribution - often a consequence of architectural nonlinearities. These small eigenvalues enlarge the volume of $\Theta_0$ and manifest as flat directions in the loss landscape. In the next section (Section~\ref{sec:spectrumbasedregretbounds}) we state explicit regret bounds based on this spectral point of view.
\end{enumerate}

\section{Spectrum-Based Regret Bounds}
\label{sec:spectrumbasedregretbounds}
The specific statement below applies to the {\em realizable case of supervised learning}, where $P(\mathcal{S}) = P(x^n, y^n) = \prod_{t=1}^n P_X(x_t)P_{\theta_0}(y_t \mymid x_t)$.
The proof of Theorem~\ref{the:supervised_local_properties_theorem} can be found in Appendix~\ref{sec:supervised_proof}. The extension to the online case can be found in Theorem~\ref{the:online_local_properties} in Appendix~\ref{sec:auxiliary_complexity_measures_online_details_realizeable} and for the batch prediction see Section~\ref{sec:auxiliary_complexity_measures_batch_details_realizeable}. The agnostic case is discussed in Appendix~\ref{sec:auxiliary_complexity_measures_details_agnostic}, see Theorem~\ref{ausilary_online_agnostic_theorem} for the online case and Theorem~\ref{the:supervisedagnostic} for the supervised case.
\begin{theorem}
\label{the:supervised_local_properties_theorem}
Assume that $\Theta \in {\mathbb R}^d$, $\|\Theta\|_2 \leq R$, with a uniform prior $w(\theta)$ over $\Theta$. Suppose that for all $\theta \in \Theta$ and all $(x^n,y^n)\in \mathcal{X}^n\times\mathcal{Y}^n$, $P_{\theta}(y^n \mymid x^n)=\prod_{t=1}^n P_{\theta}(y_t \mymid x_t)$.
Let
\[
I(\theta) 
   = \mathbbm{E}_{(x,y) \sim P_{\theta_0}(\cdot \mymid \cdot) P_{X}(\cdot) }\!\left[(\nabla_{\theta} \log P_{\theta}(y \mymid x))(\nabla_{\theta} \log P_{\theta}(y \mymid x))^\top\right].
\]
Assume that $\hat{\theta}(\mathcal{S})$, the maximum likelihood estimator,  satisfies $I(\hat{\theta}(\mathcal{S})) \approx I(\theta_0 )$ almost surely, and that Laplace’s approximation to the posterior holds. If the eigenvalues of $\mathbbm{E}_{x \sim P_X}\!\left[I(\theta_0)\right]$, denoted $\lambda_1 \ge \lambda_2 \ge \cdots \ge \lambda_d > 0$, satisfy $\lambda_{k+1} \le \frac{\alpha}{2R^2}$
for some $k \ll d,n$ and $\alpha>0$, then
\begin{align}
\label{equ:theorem1}
\regret^s_r(Q,\theta_0) \le \frac{k}{2n} + \alpha + o\!\left(\frac{1}{n}\right).
\end{align}
\end{theorem}
The proof relies on the local expansion of the KL-divergence using the FIM, bounding the contribution of the $d-k$ directions corresponding to small eigenvalues to the divergence, and using Laplace's approximation for the rest of the directions. 

When $\alpha \ll \tfrac{k}{2n}$, this simplifies to $\regret^s_r(Q,\theta_0) \le \tfrac{k}{2n} + o\!\left(\tfrac{1}{n}\right)$. In this regime, $k$ plays the role of an \emph{effective dimension}. In contrast, without the small-eigenvalue condition on the FIM, one obtains a regret of order $\tfrac{d}{2n} + o\!\left(\tfrac{1}{n}\right)$; see, for example, Theorem 1 in \cite{Aslan2006AsymptoticallyMinmaxBayes}.

In Section~\ref{app:examples_and_experiments} we present numerical experiments that illustrate the relationship between model complexity, the FIM spectrum, and the regret.



\section{The Role of the Architecture}
\label{sec:theroleofthearchitecture}
Our non-uniform regret bounds indicate that useful model classes possess a \emph{broad range of complexities}. For this range to be practically relevant, the complexities within the model class must be \emph{aligned} with the signals of interest, i.e., the signals we wish to learn should have low complexities. This raises the fundamental question of what structural properties such signals typically exhibit.

Moreover, in many scenarios the model class is not explicitly specified, but rather arises implicitly through an \emph{architecture} (e.g., a neural network defined by its width, depth, and nonlinearities, or a linear model determined by its basis functions). It is therefore essential to understand which types of architecture provide a broad range of complexities, and, given that broad range, which functions are favored by typical architectures. This inductive bias should ideally align with the characteristic properties of the signals of interest.

\subsection{Matching to Signals of Interest}
In many applications we aim to learn ``natural’’ signals (visual, auditory, or physical) and would like these to have low complexity under the chosen architecture. Because there is no universally accepted formal model of natural signals, designing architectures that assign them low complexity is part science and part art. Still, several properties are well documented to be associated with natural signals:
\begin{itemize}
\setlength{\itemsep}{-0.1cm}
    \item \emph{Smoothness}: solutions to physical PDEs are typically smooth \cite{Rao2020}.
    \item \emph{Sparsity}: natural images admit sparse representations \cite{OlshausenField1996, simoncelli2001natural}.
    \item \emph{Compositionality}: many signals and tasks exhibit hierarchical structure \cite{mhaskar2016deepvsshallownetworks, poggio2017why}.
\end{itemize}
Modern architectures align well with these desiderata:
\begin{itemize}
\setlength{\itemsep}{-0.1cm}
    \item \emph{Smoothness}: deep ReLU networks approximate Lipschitz/Sobolev functions efficiently \cite{yarotsky2017errorboundsapproximationsdeep} and display spectral bias \cite{rahaman2019spectral}.
    \item \emph{Sparsity}: $\ell_1$ regularization \cite{tibshirani1996regression} and sparse priors \cite{theodoridis2012sparsity} improve generalization; architectural sparsity boosts efficiency \cite{fedus2022review}.
    \item \emph{Compositionality}: depth yields exponential efficiency gains for certain structured functions \cite{daniely2017depthseparationneuralnetworks, devore2020neuralnetworkapproximation}.
\end{itemize}

\subsection{Large Dynamic Range: Parameter Space vs.\ Function Space} 
One potentially important source of {\em inductive bias} lies in the mapping between the parameter space and the functions space. Let $\Theta$ be the parameter space and let $\mathcal{P}\subseteq\mathcal{D}$ denote the \emph{function space} (the set of realized distributions).  
In some classical architectures (e.g., linear regression) the mapping $\theta \mapsto P_\theta$ is injective, whereas in modern architectures (DNNs, LLMs) it is highly non-injective. From an optimization perspective, non-injectivity can appear pathological and something to ``fix’’ (see \cite{neyshabur2016datadependentpathnormalizationneural}); through the lens of our regret bounds it is instead a \emph{feature}: it concentrates prior mass on particular functions and thereby induces a broad range of complexities. Moreover, our bounds account not only for the mass of a single function but also for the mass of functions that are \emph{similar} in function space, further broadening the attainable complexity range.

For example, in a DNN where only a few neurons are effectively active, the parameters associated with inactive neurons may vary over large regions without changing the output. This invariance grants the realized function substantial prior mass, see Appendix~\ref{app:zero_function_has_small_complexity}.

\subsection{Large Dynamic Range: Linear versus Layered Linear Models}

A further central source of inductive bias in modern architectures such as deep neural networks is their \emph{compositional} structure: many layers acting in sequence. We illustrate this effect first in linear settings, then in general layered functions.  

\noindent{\bf Linear Maps:}\; 
In a simple linear regression model $y = x^\top \theta + n, x\sim P_X(\cdot), n \sim \mathcal{N}(0,\sigma^2)$
the FIM reduces to the covariance of the inputs (see Appendix~\ref{app:linear_fim} for derivation). It depends only on the data distribution, not on parameters, and there is no inherent reason to expect strong spectral skew. Thus, in the purely linear case, the parameterization introduces no particular bias.

\noindent{\bf Composed Linear Maps:}\; 
Represent a linear map as a product of $\ell$ matrices, 
\[
y = A_1 A_2 \cdots A_{\ell} x + n, \;\;\; n \sim \mathcal{N}.
\]
Functionally this is equivalent to a map with a single matrix $A$, but the parameterization is different: many more factorizations $(A_1,\dots,A_{\ell})$ correspond to simple operators (e.g.\ with small or vanishing eigenvalues). 

There are several paths to see that such a structure induces a bias. First, results from random matrix theory show that products of random matrices concentrate their spectrum near zero as ${\ell}$ grows \citep{GoetzeTikhomirov2011,BougerolLacroix2014}, indicating a strong bias toward low-complexity maps. Full details and spectral density formulas are given in Appendix~\ref{app:random_products}.  
Second, one can compute the FIM for this case and show that the eigenvalues of this matrix will decrease as ${\ell}$ increases. A complete explanation is given in Appendix~\ref{app:fim_composedlinearmaps}.
Both of those analyses illustrate that layering - even when no new functions are added - changes the implicit prior over hypotheses.

\subsection{Large Dynamic Range: Nonlinear Layered Models}

The same reasoning extends to nonlinear networks. Consider a layered model
\begin{align}
\label{equ:compositestructure}
y = f_{\theta_1}(f_{\theta_2}(\cdots f_{\theta_\ell}(x)\cdots)) + n,
\quad n\sim \mathcal{N}(0,\sigma^2 I).
\end{align}
By the chain rule, gradients with respect to the parameters of layer $i$ involve products of Jacobians from earlier layers. These products tend to be degenerate, so the blocks of the FIM inherit the same spectrum-shrinking effect observed in the linear case (see Appendix~\ref{app:layered_fim} for the explicit block-matrix form).  

Intuitively, deep compositional structures therefore induce an implicit bias toward simpler effective functions - even though the class of realizable functions is unchanged. Analyses of specific cases such as ReLU networks confirm this picture \citep{KarakidaEtAl2021,Sun2025GeometricModeling}.  

Finally, the additive Gaussian assumption (\ref{equ:compositestructure}) for the model class is not essential: a layered model defined as a chain of conditional distributions $P_{\theta_i}(h_i \mymid h_{i-1})$ exhibits the same phenomenon, since the Fisher Information continues to depend on accumulated Jacobian products.  

Thus, compositional architectures naturally realize a broad dynamic range of complexities, with implicit preference for structured, low-complexity functions.

\subsection{Alternative Architectures}
Possible alternatives (e.g., finite state-space models, state-space sequence models, kernel machines, or wavelet-based constructions) can also realize a broad complexity range and good signal matching by concentrating prior mass on structured function families. Analyzing how these architectures induce complexity ranges - and how those ranges align with smoothness, sparsity, and compositionality - remains a promising direction for further study.

\section{Efficient Learning in Practice}

While the Bayesian mixture predictor is theoretically optimal in our framework, directly computing it for modern architectures is computationally infeasible.  
The mixture
\[
Q(\cdot) = \int_{\Theta} w(\theta \mid \mathcal{S}) P_{\theta}(\cdot) \, d\theta
\]
involves integrating over a parameter space $\Theta$ that, in the case of deep neural networks or transformers, may have billions of dimensions.  
Even representing the posterior $w(\theta \mid \mathcal{S})$ explicitly is not practical in such high-dimensional settings.

Despite this apparent obstacle, modern optimization algorithms - most prominently \emph{stochastic gradient descent} (SGD) and its variants - turn out to be highly effective \emph{implicit} approximators of the Bayesian mixture.  
In fact, much of the empirical success of SGD can be interpreted through this lens.

\paragraph{SGD as approximate Bayesian inference.}
From a probabilistic perspective, SGD does more than simply find a point estimate of the parameters.  
Due to the stochasticity introduced by minibatch sampling, SGD explores a region of the parameter space rather than collapsing to a single deterministic point.  
This noise, combined with the structure of the loss landscape, induces a distribution over parameters that often concentrates near regions of high posterior density.  
When SGD is augmented with \emph{Langevin dynamics} (SGD+LD) \cite{TehEtAl2015, ChenDingCarin2016}, the update rule can be seen as a discretized stochastic differential equation whose stationary distribution is exactly the Bayesian posterior:
\[
w(\theta \mid \mathcal{S}) \propto w(\theta) \, e^{-L(\theta; \mathcal{S})}.
\]
In this view, each run of SGD+LD yields a single draw $\theta$ from the posterior, and multiple independent runs approximate samples from the full mixture.

\paragraph{Practical approximations of the mixture.}
There are two main ways to exploit this connection in practice:
\begin{enumerate}
\setlength{\itemsep}{-0.1cm}
    \item \textbf{Model averaging (ensembles):} Run the training procedure multiple times with different random seeds (or inject additional noise during training), producing a set of parameter vectors $\{\theta^{(i)}\}$. The predictive distribution is then approximated by the empirical average:
    \[
    \hat{Q}(\cdot) = \frac{1}{M} \sum_{i=1}^M P_{\theta^{(i)}}(\cdot),
    \]
    where $M$ is the number of runs.  
    This is a direct finite-sample approximation to the Bayesian mixture and is closely related to classical ensemble methods \cite{SagiRokach2018}.

    \item If the training set size $n$ is large, models that differ considerably from $\theta_0$ in terms of KL-divergence will be assigned a very small weight. Thus, a single draw from SGD - often the \emph{maximum a posteriori} (MAP) estimate - is already a strong approximation to the mixture.  
    Techniques like \emph{stochastic weight averaging} (SWA) and \emph{Polyak averaging} can further refine this approximation by averaging weights along an SGD trajectory, thereby reducing variance and improving generalization without the cost of training multiple independent models.
    
\end{enumerate}

\paragraph{Scalability to high-dimensional settings.}
A key advantage of SGD-based approximations is that they scale gracefully to extremely large parameter spaces, as encountered in state-of-the-art language models and vision architectures.  
The computational cost grows roughly linearly with the number of parameters and training samples, making it feasible to apply these ideas in practice to models with billions of parameters.

\section{Experiments}
\label{app:examples_and_experiments}

Substantial empirical evidence indicates that the Fisher information matrix and the Hessian of trained models in modern architectures exhibit highly skewed spectra with many near-zero directions \cite{SagunEtAl2017,YaoEtAl2020,SankarEtAl2020}. To complement these findings, we conduct a simple experiment inspired by \cite{zhang2016understanding}, which demonstrated that even under random labels or random images that preclude generalization, networks nonetheless achieve near-perfect training accuracy, similar to the case with true labels. Since the optimization procedure (SGD without explicit regularization) is identical in both scenarios, this raises a fundamental question: why does the same training algorithm, applied to the same architecture and achieving comparable training performance, sometimes generalize and sometimes fail? We address this question through the lens of our framework.

\begin{figure}[ht]
\begin{center}
\includegraphics[width=9cm]{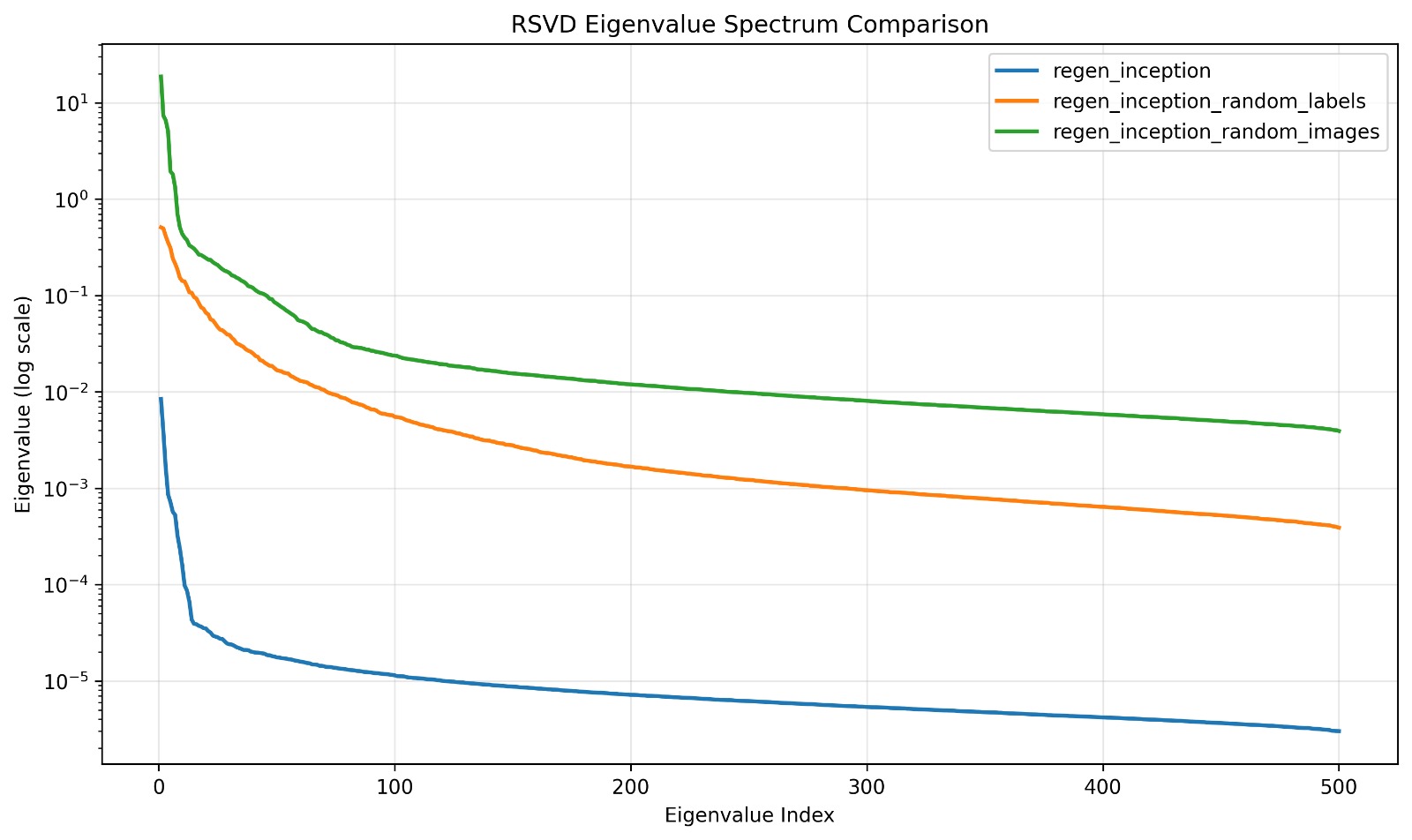}
\end{center}
\caption{Largest Eigenvalues of the empirical average Hessian (from the largest to the smallest) \label{fig:eigenvalue_distribution}}
\end{figure}

Specifically, we follow one setup in \cite{zhang2016understanding}, using the CIFAR-10 dataset and an INCEPTION network with $1,649,402$ parameters. 
The model was trained over 1000 epochs using the Adam optimizer with a learning rate of $1e-3$. 
After training over the $50,000$ training examples, the average Hessian was computed at the point of convergence and its eigenvalues were analyzed. Calculating directly the eigenvalues of such a large matrix is infeasible. Thus, the Randomized Singular Value Decomposition (RSVD) algorithm proposed by \cite{RokhlinSzlamTygert2010}, or more precisely the improved version by \cite{palitta2025rowawarerandomizedsvdapplications}, was employed to efficiently estimate the $500$ largest eigenvalues. As shown in Figure~\ref{fig:eigenvalue_distribution}, training with true labels, which generalize well, produces a highly degenerate spectrum: the eigenvalues decay rapidly, suggesting that the algorithm converges to a relatively ``simple'' model. In contrast, when trained on noisy labels or noisy images, the spectrum decays much more slowly, reflecting a ``complex'' model that fails to generalize.

\section{Our Contributions and Relation to Prior Work}
\label{sec:contributions_and_prior_work}

\paragraph{Summary of the framework.}
We develop a structured, information-theoretic framework for learning that treats training as \emph{universal prediction under log loss}. The Bayes-optimal learner in this setting is a mixture over models, and generalization is controlled by a \emph{non-uniform, model-specific complexity} determined by the posterior mass assigned to models that are close (in KL-divergence sense) to the best-in-class hypothesis. Because this true complexity can be difficult to evaluate directly, we connect it to spectral properties of the Hessian/Fisher Information Matrix (FIM), yielding tractable proxies that match phenomena observed in modern architectures and training procedures. This perspective applies uniformly across supervised, online, and batch regimes, in both realizable and agnostic (and even individual) settings.\footnote{See Secs.~2--6 for the mixture formulation, regret bounds, and complexity; Sec.~7 for architectural implications; Sec.~8 for the training connection; and Sec.~9 for empirical evidence.}

\vspace{0.25em}
\noindent\textbf{Main contributions:}
\begin{enumerate}
\setlength{\itemsep}{-0.1cm}
    \item \textbf{Unified information-theoretic framework.} We cast learning as universal prediction under log loss and analyze the Bayes mixture predictor. The resulting regret bounds are non-uniform and hold across supervised, online, and batch settings, in both realizable and agnostic cases. This places standard machine learning (ML) tasks under a single probabilistic umbrella and makes precise the role of mixture learners in high-dimensional models. [\cite{MerhavFeder1998,CoverThomas2012}] 

    \item \textbf{Non-uniform, model-specific complexity.} We introduce a complexity measure
    \(
    \mathrm{Comp}(P, {\epsilon^2}) = -  \mathbbm{E}_{{\mathcal S}}\left[\log w(\Theta_0^{\epsilon^2} \mid \mathcal{S}) \right],
    \)
    i.e., the negative log posterior mass in an $\epsilon$–ball around the best model. This quantity is \emph{data-dependent} (shrinks with more data) and \emph{architecture-aware} (captures inductive bias via prior mass). Architectures with a broad complexity range of prior complexities---as in DNNs and transformers---can be both highly expressive and still yield tight, non-uniform bounds; in contrast, classical linear models exhibit little complexity range. [\cite{GneitingRaftery2007,MerhavFeder1998}] 

    \item \textbf{Spectral (FIM/Hessian) proxies and explicit bounds.} We relate the true complexity to auxiliary measures based on the spectrum of the FIM/Hessian. Under standard conditions, if only $k \ll d$ eigenvalues are appreciable and the rest are small, our bound scales as
    \(
       R \lesssim \tfrac{k}{2n} + \alpha + o(n^{-1}),
    \)
    revealing that generalization depends on the \emph{effective} dimension captured by large eigenvalues rather than parameter count. This clarifies the connection between ``flat minima'' and low regret: flat directions correspond to large volume in parameter space near $P_{\theta_0}$, hence low complexity. [\cite{HochreiterSchmidhuber1997}] 

    \item \textbf{Architectures and inductive bias from first principles.} We show how both the non-injectivity of the  map from parameters space to functions space as well as the layered compositions bias the function space toward simpler hypotheses. For deep linear stacks, products of (nearly) independent weight matrices yield degenerate spectra, favoring few large singular values; thus, a uniform prior over parameters induces a non-uniform prior over functions with an implicit simplicity bias. This matches theoretical and empirical observations on Hessian/FIM spectra in modern networks. [\cite{AroraEtAl2019,GoetzeTikhomirov2011,BougerolLacroix2014,SagunEtAl2017,YaoEtAl2020,KarakidaEtAl2019,KarakidaEtAl2021,SankarEtAl2020,YangEtAl2022}] 
    
 \item \textbf{Layering and Composition.} Enhancing the understanding of what makes a ``good'' architecture, we observe that layering and composition in the architecture design - as in DNN's but can also be in other architectures - lead to {\em broad complexity range}. This is a general phenomenon, not unique to structures like DNN's and it is related to many works on the behavior of random matrix multiplication [\cite{GoetzeTikhomirov2011,BougerolLacroix2014}]. DNN's may fit tasks like image recognition, but for tasks like language models or speech recognition other architectures that have also the composition of functions property - such as cascaded finite state machines, may be more natural. 

    \item \textbf{Practical learners as approximate mixtures.} We interpret SGD (and SGD with Langevin noise) as scalable approximations to Bayesian mixtures, explaining why single solutions, SWA/ensembles, and posterior sampling-like procedures work well in practice despite the intractability of exact mixing in large architectures. [\cite{TehEtAl2015,ChenDingCarin2016,SagiRokach2018}] 

    \item \textbf{Coding/compression interpretation.} Our regret bounds admit a codelength view: $-\log w(\Theta_0)$ is the \emph{description length} of a model and its $\epsilon$–neighborhood, providing an operational link to MDL and compression-style generalization arguments while avoiding pre-specified compressed classes. [\cite{Rissanen1984,HintonVanCamp1993,AroraEtAl2018}] 

\end{enumerate}

\paragraph{Relation to prior work.}
\emph{Universal prediction and non-uniform bounds.}
Our online regret bound in the realizable case specializes to Barron’s Bayesian “index of resolvability,” while our presentation targets the agnostic setting and emphasizes non-uniformity and architectural bias. [\cite{Barron1987,Barron1988}] We also connect to classical universal coding and to modern universal prediction (batch/online) formulations. [\cite{MerhavFeder1998,BondaschiGastpar2024,FogelFeder2017,ClarkeBarron1994}] 

\emph{PAC-Bayes.}
Structurally, both frameworks reason about priors/posteriors and involve KL terms, but the divergence enters differently: PAC-Bayes controls $D_{\text{KL}}(\text{posterior}\,\|\,\text{prior})$ and relies critically on prior design, whereas our log-loss regret naturally yields $D_{\text{KL}}(P_{\theta_0}\,\|\,Q)$ and places less emphasis on the specific prior choice. Recent work that links PAC-Bayes to flat minima is complementary to our FIM-based view. [\cite{McAllester1999,DziugaiteRoy2017,Alquier2021,HaddoucheEtAl2025}] 

\emph{Flat minima, sharpness, and spectra.}
The empirical success of flat solutions has a long history; our analysis makes precise \emph{why} flatness can correlate with generalization in log-loss: flat regions imply large posterior mass near $P_{\theta_0}$. [\cite{HochreiterSchmidhuber1997,HintonVanCamp1993}] Related algorithmic advances (e.g., SAM) and follow-ups show nuanced relationships between sharpness measures and generalization, in line with our spectral viewpoint. [\cite{ForetEtAl2021,WenLiMa2023,KuzborskijEtAl2019}] Moreover, large-scale studies consistently report many near-zero FIM/Hessian eigenvalues in trained networks, matching our predictions; our MNIST experiment exhibits the same pattern. [\cite{SagunEtAl2017,YaoEtAl2020,KarakidaEtAl2019,KarakidaEtAl2021,SankarEtAl2020,YangEtAl2022}] 

\emph{Singular Learning Theory.}
A prior work that is highly relevant to our results is singular learning theory (SLT), see \cite{watanabe2010asymptotic}, \cite{watanabe2024recent} and references therein. This framework utilizes the Resolution Theorem by \cite{hironaka1964resolution} to characterize the asymptotic values of the regret for the online and batch learning problem for memoryless hypothesis classes, both in the realizable and the agnostic setting. This highly relevant theory examines the asymptotic results while we have non-asymptotic upper bounds, which also hold for hypothesis classes and data-generating distributions with memory. Still,  
the interpretation of the real log canonical threshold as the $\lambda = \lim_{\epsilon \to 0} \frac{\log \left( w(\Theta_0) \right)}{\log(\epsilon^2)}$, can be easily incorporated into the online regret bound we have in Appendix \ref{sec:regret_boundsonline}, to get a result which is equivalent to Theorem \ref{the:online_local_properties}.

\emph{Compression, distillation, and Occam.}
Compression-style generalization bounds and model distillation echo the codelength interpretation of our regret; our framework recovers this view directly from universal prediction without pre-defining compressed classes. Recent evidence that DNNs have an “inbuilt Occam’s razor” complements our non-uniform complexity picture. [\cite{AroraEtAl2018,HsiehEtAl2023,MingardEtAl2025,GluchUrbanke2023}] 

\paragraph{Scope and limitations.}
Our spectral bound currently depends on the input distribution, e.g., via the definition of $I(\theta_0)$ in Theorem~\ref{the:supervised_local_properties_theorem}. Developing sample-based variants and broadening the analysis beyond local Laplace approximations are natural next steps. We view our small-scale experiments as illustrative rather than exhaustive and expect larger studies to further probe when architectures exhibit a large complexity range matched to natural signals. 

\newpage

\vskip 0.2in

\bibliographystyle{unsrtnat}
\bibliography{references}  

\begin{thebibliography}{65}
\providecommand{\natexlab}[1]{#1}
\providecommand{\url}[1]{\texttt{#1}}
\expandafter\ifx\csname urlstyle\endcsname\relax
  \providecommand{\doi}[1]{doi: #1}\else
  \providecommand{\doi}{doi: \begingroup \urlstyle{rm}\Url}\fi

\bibitem[Gneiting and Raftery(2007)]{GneitingRaftery2007}
Tilmann Gneiting and Adrian~E Raftery.
\newblock Strictly proper scoring rules, prediction, and estimation.
\newblock \emph{Journal of the American Statistical Association}, 102\penalty0 (477):\penalty0 359--378, 2007.

\bibitem[Casella and Berger(2002)]{casella2002statistical}
George Casella and Roger~L Berger.
\newblock \emph{Statistical Inference}.
\newblock Duxbury Press, 2nd edition, 2002.

\bibitem[Merhav and Feder(1998)]{MerhavFeder1998}
Neri Merhav and Meir Feder.
\newblock Universal prediction.
\newblock \emph{IEEE Transactions on Information Theory}, 44\penalty0 (6):\penalty0 2124--2147, 1998.
\newblock \doi{10.1109/18.720534}.

\bibitem[Cover and Thomas(2012)]{CoverThomas2012}
Thomas~M Cover and Joy~A Thomas.
\newblock \emph{Elements of information theory}.
\newblock John Wiley \& Sons, 2012.

\bibitem[Niculescu-Mizil and Caruana(2005)]{niculescu2005predicting}
Alexandru Niculescu-Mizil and Rich Caruana.
\newblock Predicting good probabilities with supervised learning.
\newblock In \emph{Proceedings of the 22nd International Conference on Machine Learning (ICML)}, pages 625--632. ACM, 2005.

\bibitem[Gluch and Urbanke(2023)]{GluchUrbanke2023}
Grzegorz Gluch and R{\"{u}}diger~L. Urbanke.
\newblock Bayes complexity of learners vs overfitting.
\newblock \emph{CoRR}, abs/2303.07874, 2023.
\newblock URL \url{https://doi.org/10.48550/arXiv.2303.07874}.

\bibitem[Hochreiter and Schmidhuber(1997)]{HochreiterSchmidhuber1997}
Sepp Hochreiter and J\"{u}rgen Schmidhuber.
\newblock Flat minima.
\newblock \emph{Neural Comput.}, 9\penalty0 (1):\penalty0 1–42, January 1997.
\newblock ISSN 0899-7667.
\newblock URL \url{https://doi.org/10.1162/neco.1997.9.1.1}.

\bibitem[Aslan(2006)]{Aslan2006AsymptoticallyMinmaxBayes}
Mihaela Aslan.
\newblock {Asymptotically minimax Bayes predictive densities}.
\newblock \emph{The Annals of Statistics}, 34\penalty0 (6):\penalty0 2921 -- 2938, 2006.
\newblock \doi{10.1214/009053606000000885}.
\newblock URL \url{https://doi.org/10.1214/009053606000000885}.

\bibitem[Rao et~al.(2020)Rao, Sun, and Liu]{Rao2020}
Chengping Rao, Hao Sun, and Yang Liu.
\newblock Physics-informed deep learning for incompressible laminar flows.
\newblock \emph{Theoretical and Applied Mechanics Letters}, 10\penalty0 (3):\penalty0 207--212, 2020.
\newblock \doi{10.1016/j.taml.2020.01.039}.

\bibitem[Olshausen and Field(1996)]{OlshausenField1996}
Bruno~A. Olshausen and David~J. Field.
\newblock Emergence of simple-cell receptive field properties by learning a sparse code for natural images.
\newblock \emph{Nature}, 381\penalty0 (6583):\penalty0 607--609, June 1996.
\newblock \doi{10.1038/381607a0}.

\bibitem[Simoncelli and Freeman(2001)]{simoncelli2001natural}
Eero~P. Simoncelli and William~T. Freeman.
\newblock Natural image statistics and neural representation.
\newblock \emph{Neural Information Processing Systems (NIPS)}, 13:\penalty0 153--159, 2001.

\bibitem[Mhaskar and Poggio(2016)]{mhaskar2016deepvsshallownetworks}
Hrushikesh Mhaskar and Tomaso Poggio.
\newblock Deep vs. shallow networks : An approximation theory perspective, 2016.
\newblock URL \url{https://arxiv.org/abs/1608.03287}.

\bibitem[Poggio et~al.(2017)Poggio, Mhaskar, Rosasco, Miranda, and Liao]{poggio2017why}
Tomaso Poggio, Hrushikesh Mhaskar, Lorenzo Rosasco, Brando Miranda, and Qianli Liao.
\newblock Why and when can deep-but not shallow-networks avoid the curse of dimensionality: A review.
\newblock \emph{International Journal of Automation and Computing}, 14\penalty0 (5):\penalty0 503--519, 2017.
\newblock ISSN 1745-0647.
\newblock \doi{10.1007/s11633-017-1054-2}.
\newblock URL \url{https://www.mi-research.net/article/doi/10.1007/s11633-017-1054-2}.

\bibitem[Yarotsky(2017)]{yarotsky2017errorboundsapproximationsdeep}
Dmitry Yarotsky.
\newblock Error bounds for approximations with deep relu networks, 2017.
\newblock URL \url{https://arxiv.org/abs/1610.01145}.

\bibitem[Rahaman et~al.(2019)Rahaman, Baratin, Arpit, Draxler, Lin, Hamprecht, Bengio, and Courville]{rahaman2019spectral}
Nasim Rahaman, Aristide Baratin, Devansh Arpit, Felix Draxler, Min Lin, Fred~A. Hamprecht, Yoshua Bengio, and Aaron Courville.
\newblock On the spectral bias of neural networks.
\newblock In \emph{Proceedings of the 36th International Conference on Machine Learning (ICML)}, volume~97, pages 5301--5310. PMLR, 2019.
\newblock URL \url{https://proceedings.mlr.press/v97/rahaman19a.html}.

\bibitem[Tibshirani(1996)]{tibshirani1996regression}
Robert Tibshirani.
\newblock Regression shrinkage and selection via the lasso.
\newblock \emph{Journal of the Royal Statistical Society: Series B}, 58\penalty0 (1):\penalty0 267--288, 1996.
\newblock \doi{10.1111/j.2517-6161.1996.tb02080.x}.

\bibitem[Theodoridis et~al.(2012)Theodoridis, Kopsinis, and Slavakis]{theodoridis2012sparsity}
Sergios Theodoridis, Yannis Kopsinis, and Konstantinos Slavakis.
\newblock Sparsity-aware learning and compressed sensing: An overview.
\newblock \emph{arXiv preprint arXiv:1211.5231}, 2012.
\newblock URL \url{https://arxiv.org/abs/1211.5231}.

\bibitem[Fedus et~al.(2022)Fedus, Dean, and Zoph]{fedus2022review}
William Fedus, Jeff Dean, and Barret Zoph.
\newblock A review of sparse expert models in deep learning.
\newblock \emph{arXiv preprint arXiv:2209.01667}, 2022.
\newblock URL \url{https://arxiv.org/abs/2209.01667}.

\bibitem[Daniely(2017)]{daniely2017depthseparationneuralnetworks}
Amit Daniely.
\newblock Depth separation for neural networks, 2017.
\newblock URL \url{https://arxiv.org/abs/1702.08489}.

\bibitem[DeVore et~al.(2020)DeVore, Hanin, and Petrova]{devore2020neuralnetworkapproximation}
Ronald DeVore, Boris Hanin, and Guergana Petrova.
\newblock Neural network approximation, 2020.
\newblock URL \url{https://arxiv.org/abs/2012.14501}.

\bibitem[Neyshabur et~al.(2016)Neyshabur, Tomioka, Salakhutdinov, and Srebro]{neyshabur2016datadependentpathnormalizationneural}
Behnam Neyshabur, Ryota Tomioka, Ruslan Salakhutdinov, and Nathan Srebro.
\newblock Data-dependent path normalization in neural networks, 2016.
\newblock URL \url{https://arxiv.org/abs/1511.06747}.

\bibitem[Götze and Tikhomirov(2011)]{GoetzeTikhomirov2011}
Friedrich Götze and Alexander Tikhomirov.
\newblock On the asymptotic spectrum of products of independent random matrices, 2011.
\newblock URL \url{https://arxiv.org/abs/1012.2710}.

\bibitem[Bougerol and Lacroix(2014)]{BougerolLacroix2014}
Philippe Bougerol and Jean Lacroix.
\newblock \emph{Products of Random Matrices with Applications to Schrodinger Operators}.
\newblock Springer, 2014.
\newblock ISBN 9781468491739.
\newblock URL \url{https://books.google.co.il/books?id=u3kPswEACAAJ}.

\bibitem[Karakida et~al.(2021)Karakida, Akaho, and Amari]{KarakidaEtAl2021}
Ryo Karakida, Shotaro Akaho, and Shun-ichi Amari.
\newblock Pathological spectra of the {F}isher information metric and its variants in deep neural networks.
\newblock \emph{Neural Computation}, 33\penalty0 (8):\penalty0 2274--2307, 07 2021.
\newblock URL \url{https://doi.org/10.1162/neco\_a\_01411}.

\bibitem[Sun and Nielsen(2025)]{Sun2025GeometricModeling}
Ke~Sun and Frank Nielsen.
\newblock A geometric modeling of occam’s razor in deep learning.
\newblock \emph{Information Geometry}, June 2025.
\newblock \doi{10.1007/s41884-025-00167-2}.
\newblock URL \url{https://Vinyals.springer.com/article/10.1007/s41884-025-00167-2}.

\bibitem[Teh et~al.(2015)Teh, Thiéry, and Vollmer]{TehEtAl2015}
Yee~Whye Teh, Alexandre Thiéry, and Sebastian Vollmer.
\newblock Consistency and fluctuations for stochastic gradient langevin dynamics, 2015.

\bibitem[Chen et~al.(2016)Chen, Ding, and Carin]{ChenDingCarin2016}
Changyou Chen, Nan Ding, and Lawrence Carin.
\newblock On the convergence of stochastic gradient mcmc algorithms with high-order integrators, 2016.

\bibitem[Sagi and Rokach(2018)]{SagiRokach2018}
Omer Sagi and Lior Rokach.
\newblock Ensemble learning: A survey.
\newblock \emph{WIREs Data Mining and Knowledge Discovery}, 8\penalty0 (4):\penalty0 e1249, 2018.
\newblock \doi{https://doi.org/10.1002/widm.1249}.
\newblock URL \url{https://wires.onlinelibrary.wiley.com/doi/abs/10.1002/widm.1249}.

\bibitem[Sagun et~al.(2017)Sagun, Bottou, and LeCun]{SagunEtAl2017}
Levent Sagun, Leon Bottou, and Yann LeCun.
\newblock Eigenvalues of the hessian in deep learning: Singularity and beyond, 2017.
\newblock URL \url{https://openreview.net/forum?id=B186cP9gx}.

\bibitem[Yao et~al.(2020)Yao, Gholami, Keutzer, and Mahoney]{YaoEtAl2020}
Zhewei Yao, Amir Gholami, Kurt Keutzer, and Michael~W. Mahoney.
\newblock Pyhessian: Neural networks through the lens of the hessian.
\newblock In \emph{IEEE BigData}, pages 581--590, 2020.
\newblock URL \url{https://doi.org/10.1109/BigData50022.2020.9378171}.

\bibitem[Sankar et~al.(2020)Sankar, Khasbage, Vigneswaran, and Balasubramanian]{SankarEtAl2020}
Adepu~Ravi Sankar, Yash Khasbage, Rahul Vigneswaran, and Vineeth~N Balasubramanian.
\newblock A deeper look at the hessian eigenspectrum of deep neural networks and its applications to regularization, 2020.
\newblock URL \url{https://arxiv.org/abs/2012.03801}.

\bibitem[Zhang et~al.(2016)Zhang, Bengio, Hardt, Recht, and Vinyals]{zhang2016understanding}
Chiyuan Zhang, Samy Bengio, Moritz Hardt, Benjamin Recht, and Oriol Vinyals.
\newblock Understanding deep learning requires rethinking generalization.
\newblock \emph{arXiv preprint arXiv:1611.03530}, 2016.
\newblock URL \url{https://arxiv.org/abs/1611.03530?utm_source=chatgpt.com}.

\bibitem[Rokhlin et~al.(2010)Rokhlin, Szlam, and Tygert]{RokhlinSzlamTygert2010}
Vladimir Rokhlin, Arthur Szlam, and Mark Tygert.
\newblock A randomized algorithm for principal component analysis.
\newblock \emph{SIAM Journal on Matrix Analysis and Applications}, 31\penalty0 (3):\penalty0 1100--1124, 2010.
\newblock \doi{10.1137/080736417}.

\bibitem[Palitta and Portaro(2025)]{palitta2025rowawarerandomizedsvdapplications}
Davide Palitta and Sascha Portaro.
\newblock Row-aware randomized svd with applications, 2025.
\newblock URL \url{https://arxiv.org/abs/2408.04503}.

\bibitem[Arora et~al.(2019)Arora, Cohen, Golowich, and Hu]{AroraEtAl2019}
Sanjeev Arora, Nadav Cohen, Noah Golowich, and Wei Hu.
\newblock A convergence analysis of gradient descent for deep linear neural networks.
\newblock In \emph{International Conference on Learning Representations}, 2019.
\newblock URL \url{https://openreview.net/forum?id=SkMQg3C5K7}.

\bibitem[Karakida et~al.(2019)Karakida, Akaho, and ichi Amari]{KarakidaEtAl2019}
Ryo Karakida, Shotaro Akaho, and Shun ichi Amari.
\newblock Universal statistics of {F}isher information in deep neural networks: Mean field approach, 2019.
\newblock URL \url{https://arxiv.org/abs/1806.01316}.

\bibitem[Yang et~al.(2022)Yang, Mao, and Chaudhari]{YangEtAl2022}
Rubing Yang, Jialin Mao, and Pratik Chaudhari.
\newblock Does the data induce capacity control in deep learning?, 2022.
\newblock URL \url{https://arxiv.org/abs/2110.14163}.

\bibitem[Rissanen(1984)]{Rissanen1984}
Jorma Rissanen.
\newblock Universal coding, information, prediction, and estimation§.
\newblock \emph{IEEE Transactions on Information theory}, 30\penalty0 (4):\penalty0 629--636, 1984.

\bibitem[Hinton and van Camp(1993)]{HintonVanCamp1993}
Geoffrey~E. Hinton and Drew van Camp.
\newblock Keeping the neural networks simple by minimizing the description length of the weights.
\newblock In \emph{Proceedings of the Sixth Annual Conference on Computational Learning Theory}, COLT '93, page 5–13, New York, NY, USA, 1993. Association for Computing Machinery.
\newblock ISBN 0897916115.
\newblock URL \url{https://doi.org/10.1145/168304.168306}.

\bibitem[Arora et~al.(2018)Arora, Ge, Neyshabur, and Zhang]{AroraEtAl2018}
Sanjeev Arora, Rong Ge, Behnam Neyshabur, and Yi~Zhang.
\newblock Stronger generalization bounds for deep nets via a compression approach.
\newblock \emph{arXiv preprint arXiv:1802.05296}, 2018.

\bibitem[Barron(1987)]{Barron1987}
Andrew~R. Barron.
\newblock \emph{Are Bayes rules consistent in information?}, pages 85--91.
\newblock Springer New York, New York, NY, 1987.
\newblock URL \url{https://doi.org/10.1007/978-1-4612-4808-8_22}.

\bibitem[Barron(1988)]{Barron1988}
Andrew~R. Barron.
\newblock The exponential convergence of posterior probabilities with implications for {B}ayes estimators of density functions.
\newblock Technical Report~7, University of Illinois at Urbana-Champaign, Department of Statistics, Champaign, IL, 1988.
\newblock URL \url{http://www.stat.yale.edu/~arb4/publications_files/convergence%20of%20bayer%27s%20estimator.pdf}.

\bibitem[Bondaschi and Gastpar(2024)]{BondaschiGastpar2024}
Marco Bondaschi and Michael Gastpar.
\newblock Batch universal prediction.
\newblock \emph{arXiv preprint arXiv:2402.03901}, 2024.

\bibitem[Fogel and Feder(2017)]{FogelFeder2017}
Yaniv Fogel and Meir Feder.
\newblock On the problem of on-line learning with log-loss.
\newblock In \emph{Information Theory (ISIT), 2017 IEEE International Symposium on}, pages 2995--2999. IEEE, 2017.

\bibitem[Clarke and Barron(1994)]{ClarkeBarron1994}
Bertrand~S Clarke and Andrew~R Barron.
\newblock Jeffreys' prior is asymptotically least favorable under entropy risk.
\newblock \emph{Journal of Statistical planning and Inference}, 41\penalty0 (1):\penalty0 37--60, 1994.

\bibitem[McAllester(1999)]{McAllester1999}
David~A McAllester.
\newblock {PAC-B}ayesian model averaging.
\newblock In \emph{Proceedings of the twelfth annual conference on Computational learning theory}, pages 164--170, 1999.

\bibitem[Dziugaite and Roy(2017)]{DziugaiteRoy2017}
Gintare~Karolina Dziugaite and Daniel~M Roy.
\newblock Computing nonvacuous generalization bounds for deep (stochastic) neural networks with many more parameters than training data.
\newblock \emph{arXiv preprint arXiv:1703.11008}, 2017.

\bibitem[Alquier(2024)]{Alquier2021}
Pierre Alquier.
\newblock User-friendly introduction to {PAC-B}ayes bounds.
\newblock \emph{Foundations and Trends® in Machine Learning}, 17\penalty0 (2):\penalty0 174–303, 2024.
\newblock ISSN 1935-8245.
\newblock \doi{10.1561/2200000100}.
\newblock URL \url{http://dx.doi.org/10.1561/2200000100}.

\bibitem[Haddouche et~al.(2025)Haddouche, Viallard, Simsekli, and Guedj]{HaddoucheEtAl2025}
Maxime Haddouche, Paul Viallard, Umut Simsekli, and Benjamin Guedj.
\newblock A pac-bayesian link between generalisation and flat minima, 2025.
\newblock URL \url{https://arxiv.org/abs/2402.08508}.

\bibitem[Foret et~al.(2021)Foret, Kleiner, Mobahi, and Neyshabur]{ForetEtAl2021}
Pierre Foret, Ariel Kleiner, Hossein Mobahi, and Behnam Neyshabur.
\newblock Sharpness-aware minimization for efficiently improving generalization.
\newblock In \emph{International Conference on Learning Representations (ICLR)}, April 2021.
\newblock URL \url{https://arxiv.org/abs/2010.01412}.

\bibitem[Wen et~al.(2023)Wen, Li, and Ma]{WenLiMa2023}
Kaiyue Wen, Zhiyuan Li, and Tengyu Ma.
\newblock Sharpness minimization algorithms do not only minimize sharpness to achieve better generalization.
\newblock In \emph{Advances in Neural Information Processing Systems (NeurIPS)}, volume~36, 2023.
\newblock URL \url{https://proceedings.neurips.cc/paper_files/paper/2023/hash/0354767c6386386be17cabe4fc59711b-Abstract-Conference.html}.

\bibitem[Kuzborskij et~al.(2019)Kuzborskij, Cesa-Bianchi, and Szepesv\'ari]{KuzborskijEtAl2019}
Ilja Kuzborskij, Nicol\`{o} Cesa-Bianchi, and Csaba Szepesv\'ari.
\newblock Distribution-dependent analysis of {Gibbs-ERM} principle.
\newblock In Alina Beygelzimer and Daniel Hsu, editors, \emph{Proceedings of the Thirty-Second Conference on Learning Theory}, volume~99 of \emph{Proceedings of Machine Learning Research}, pages 2028--2054. PMLR, June 2019.
\newblock URL \url{https://proceedings.mlr.press/v99/kuzborskij19a.html}.

\bibitem[Watanabe and Opper(2010)]{watanabe2010asymptotic}
Sumio Watanabe and Manfred Opper.
\newblock Asymptotic equivalence of bayes cross validation and widely applicable information criterion in singular learning theory.
\newblock \emph{Journal of machine learning research}, 11\penalty0 (12), 2010.

\bibitem[Watanabe(2024)]{watanabe2024recent}
Sumio Watanabe.
\newblock Recent advances in algebraic geometry and bayesian statistics.
\newblock \emph{Information Geometry}, 7\penalty0 (Suppl 1):\penalty0 187--209, 2024.

\bibitem[Hironaka(1964)]{hironaka1964resolution}
Heisuke Hironaka.
\newblock Resolution of singularities of an algebraic variety over a field of characteristic zero: Ii.
\newblock \emph{Annals of Mathematics}, 79\penalty0 (2):\penalty0 205--326, 1964.

\bibitem[Hsieh et~al.(2023)Hsieh, Li, Yeh, Nakhost, Fujii, Ratner, Krishna, Lee, and Pfister]{HsiehEtAl2023}
Cheng-Yu Hsieh, Chun-Liang Li, Chih-Kuan Yeh, Hootan Nakhost, Yasuhisa Fujii, Alexander Ratner, Ranjay Krishna, Chen-Yu Lee, and Tomas Pfister.
\newblock Distilling step-by-step! outperforming larger language models with less training data and smaller model sizes, 2023.
\newblock URL \url{https://arxiv.org/abs/2305.02301}.

\bibitem[Mingard et~al.(2025)Mingard, Rees, Valle-Pérez, and Louis]{MingardEtAl2025}
Chris Mingard, Henry Rees, Guillermo Valle-Pérez, and Ard Louis.
\newblock Deep neural networks have an inbuilt occam’s razor.
\newblock \emph{Nature Communications}, 16, 01 2025.
\newblock \doi{10.1038/s41467-024-54813-x}.

\bibitem[Delétang et~al.(2024)Delétang, Ruoss, Duquenne, Catt, Genewein, Mattern, Grau-Moya, Wenliang, Aitchison, Orseau, Hutter, and Veness]{delétang2024languagemodelingcompression}
Grégoire Delétang, Anian Ruoss, Paul-Ambroise Duquenne, Elliot Catt, Tim Genewein, Christopher Mattern, Jordi Grau-Moya, Li~Kevin Wenliang, Matthew Aitchison, Laurent Orseau, Marcus Hutter, and Joel Veness.
\newblock Language modeling is compression, 2024.
\newblock URL \url{https://arxiv.org/abs/2309.10668}.

\bibitem[Fogel and Feder(2025)]{Fogel2024Combined}
Yaniv Fogel and Meir Feder.
\newblock Combined batch and online universal prediction.
\newblock In \emph{Information Theory, Probability and Statistical Learning}, 2025.

\bibitem[Csisz{\'a}r(1975)]{csiszar1975}
Imre Csisz{\'a}r.
\newblock I-divergence geometry of probability distributions and minimization problems.
\newblock \emph{Ann. Probab.}, 3:\penalty0 146 -- 158, 1975.

\bibitem[Feder and Merhav(1996)]{FederMerhavHierarchical}
Meir Feder and Neri Merhav.
\newblock Hierarchical universal coding.
\newblock \emph{IEEE Transactions on Information Theory}, 42\penalty0 (5):\penalty0 1354--1364, 1996.
\newblock \doi{10.1109/18.532877}.

\bibitem[Elias(1975)]{1055349}
Peter Elias.
\newblock Universal codeword sets and representations of the integers.
\newblock \emph{IEEE Transactions on Information Theory}, 21\penalty0 (2):\penalty0 194--203, 1975.
\newblock \doi{10.1109/TIT.1975.1055349}.

\bibitem[Laurent and Massart(2000)]{laurent2000adaptive}
Beatrice Laurent and Pascal Massart.
\newblock Adaptive estimation of a quadratic functional by model selection.
\newblock \emph{Annals of statistics}, pages 1302--1338, 2000.

\bibitem[He et~al.(2016)He, Guan, and Jha]{he2016improve}
Jingjing He, Xuefei Guan, and Ratneshwar Jha.
\newblock Improve the accuracy of asymptotic approximation in reliability problems involving multimodal distributions.
\newblock \emph{IEEE Transactions on Reliability}, 65\penalty0 (4):\penalty0 1724--1736, 2016.

\bibitem[Cesa-Bianchi et~al.(2004)Cesa-Bianchi, Conconi, and Gentile]{cesa2004generalization}
Nicolo Cesa-Bianchi, Alex Conconi, and Claudio Gentile.
\newblock On the generalization ability of on-line learning algorithms.
\newblock \emph{IEEE Transactions on Information Theory}, 50\penalty0 (9):\penalty0 2050--2057, 2004.

\end{thebibliography}

\newpage

\appendix

\section{Setup, Learning as Probabilistic Prediction, and Regret -- Details} 
\label{sec:learning_details}
We cover the supervised setting in the main text. The appendices present the corresponding material for the \emph{online} and \emph{batch} cases, along with all proofs. We begin with the online case, as it is usually the most straightforward to define and analyze, and it aligns closely with the information-theoretic foundations of universal prediction. We then turn to the batch case, and finally to the supervised case.

\subsection{Language Models -- Online Case}
\label{sec:model_online}
Consider {\em language models}, starting with the {\em online case}. The model class is defined as  
\begin{align}
\Theta^o = \{P_{\theta}(x^n): x^n \in  {\mathcal X}^n;\; P_{\theta}(x^n) \geq 0;\; \sum_{x^n} P_{\theta}(x^n)=1 \}_{\theta \in \Theta}. \label{equ:model_class_online}
\end{align} 
In words, for each parameter $\theta \in \Theta$, the distribution $P_{\theta}(\cdot)$ specifies a probability law over ${\mathcal X}^n$, the set of sequences of length $n$ with entries from ${\mathcal X}$.
We denote an element of ${\mathcal X}^n$ by $x^n$, with its $t$-th entry written as $x_t$. 

Specifying $P_{\theta}(x^n)$ automatically determines the distributions $P_{\theta}(x^t)$ for $t=1,\dots,n-1$ by marginalization, as well as the conditional distributions 
\[
P_{\theta}(x_t \mymid x^{t-1}) = \frac{P_{\theta}(x^t)}{P_{\theta}(x^{t-1})}, \qquad t=2,\dots,n.
\]
The \emph{learner} ${\mathcal A}$ produces its own distribution $Q(\cdot)$ over ${\mathcal X}^n$. Just as with the model class, specifying $Q(x^n)$ is equivalent to specifying the conditional distributions $Q(x_t \mymid x^{t-1})$ for $t=1,\dots,n$, since
\[
Q(x^n) = \prod_{t=1}^n Q(x_t \mymid x^{t-1}).
\]
The sequence $x^n$ is revealed \emph{sequentially}. At time $t$, with $1 \leq t \leq n$, the learner ${\mathcal A}$ has observed $x^{t-1}$ and produces a predictive distribution for $x_t$, namely $Q(x_t \mymid x^{t-1})$. Then the actual $x_t$ is revealed. We assume the loss function is the \emph{log-loss}, so the learner incurs
\[
-\log\bigl(Q(x_t \mymid x^{t-1})\bigr).
\]
Assume that the data is generated by some distribution $P(x^n)$, and hence by $P(x_t \mymid x^{t-1})$ at each step. This distribution need not belong to the model class. Let $P_{\theta_0}$ denote the model in the class closest to $P$ in KL divergence:
\begin{align}
\theta_0 = \arg \min_{\theta} D_{\text{KL}}\bigl(P(x^n) \mydoublemid P_{\theta}(x^n)\bigr).
\label{equ:theta0_online}
\end{align}
The \emph{pointwise regret} of predicting with $Q(x_t \mymid x^{t-1})$ instead of with the best model $P_{\theta_0}(x_t \mymid x^{t-1})$ is
\[
-\log Q(x_t \mymid x^{t-1}) + \log P_{\theta_0}(x_t \mymid x^{t-1})
= \log \frac{P_{\theta_0}(x_t \mymid x^{t-1})}{Q(x_t \mymid x^{t-1})}.
\]
Summing over $n$ steps, the total regret is
\begin{align*}
\sum_{t=1}^{n} \log \frac{P_{\theta_0}(x_t \mymid x^{t-1})}{Q(x_t \mymid x^{t-1})}
= \log \frac{P_{\theta_0}(x^n)}{Q(x^n)}.
\end{align*}
The {\em expected regret}, normalized by sequence length, is
\begin{align}
\regret_a^o(Q,P) 
&= \frac{1}{n} \mathbbm{E}_{x^n \sim P(x^n)}\Bigl[ \log \frac{P_{\theta_0}(x^n)}{Q(x^n)} \Bigr]\label{equ:regret_online} \\
&= \underbrace{\tfrac{1}{n} D\bigl(P_{\theta_0}(X^n)\mydoublemid Q(X^n)\bigr)}_{\text{if $P(x^n)=P_{\theta_0}(x^n)$, i.e., if realizable}}. \notag
\end{align}
This setup is known as {\em universal prediction} \citet{MerhavFeder1998}. In practice, it captures the core task of large language models (LLMs): predicting the next token given past context - see \citet{delétang2024languagemodelingcompression} for a detailed discussion. The fully online setting is somewhat atypical for LLMs, since it assumes no pre-training. In the next section we turn to the {\em batch setting}, where a fixed initial sequence is provided for training before prediction begins. Although other variants exist - some closer to how modern LLMs are actually trained (e.g., \citet{BondaschiGastpar2024,Fogel2024Combined}) - we restrict ourselves to these two cases for clarity of exposition.

\subsection{Language Models -- Batch Cases}
\label{sec:model_batch}
We stay with {\em language models} but we turn our attention to the {\em batch} case. Our model class remains $\Theta^b=\Theta^o$, see (\ref{equ:model_class_online}). Of particular importance for the batch case is the \emph{conditional distribution} $P_{\theta}(x_{n} \mymid x^{n-1})= \frac{P_{\theta}(x^{n})}{P_{\theta}(x^{n-1})}$.

Given $x^{n-1}$, the learner ${\mathcal A}$ is asked to produce a probability distribution on $x_{n}$. Denote this distribution by $Q(x_{n} \mymid x^{n-1})$. The actual $x_{n}$ is then revealed. We assume again that the \emph{loss} function is the \emph{log-loss}. That is,  the learner incurs a loss of
\[
-\log\bigl(Q(x_{n} \mymid x^{n-1})\bigr).
\]
Assume that the data is generated by some distribution $P(x^n)$, and hence by $P(x_n \mymid x^{n-1})$ at the prediction step. As for the online case, this distribution need not belong to the model class. Let $P_{\theta_0}$ denote the model in the model class closest to $P$ in KL divergence as specified in (\ref{equ:theta0_online}).
The \emph{pointwise regret} of using $Q(x_{n} \mymid x^{n-1})$ instead of the true model $P_{\theta_0}(x_{n} \mymid x^{n-1})$ is
\[
-\log\bigl(Q(x_{n} \mymid x^{n-1})\bigr)\;-\;\bigl(-\log\bigl(P_{\theta_0}(x_{n} \mymid x^{n-1})\bigr)\bigr)
\;=\;\log\frac{P_{\theta_0}(x_{n} \mymid x^{n-1})}{Q(x_{n} \mymid x^{n-1})}.
\]
The {\em expected regret} is then
\begin{align}
\label{equ:regret_batch}
    \regret_a^b(Q,P) & = 
   {\mathbbm E}_{x^n \sim P_{\theta_0}} \Bigl[
   \log \left( \frac{P_{\theta_0}(x_n\mymid x^{n-1})}{Q(x_n\mymid x^{n-1})} \right) \Bigr]  \\
   = & \underbrace{D_{\text{KL}}\left(P_{\theta_0}(x_n \mymid x^{n-1})\mydoublemid Q(x_n \mymid  x^{n-1}) | X^{n-1} \right).}_{\text{if $P(x^n)=P_{\theta_0}(x^n)$, i.e., if realizable}} \notag
\end{align}

Note that, although not explicitly specified, the expected regret (of both the online and the batch) depends on the data size $n$. 

\section{Mixture Models -- Details}
\label{sec:mixture_modelsdetails}
In (\ref{equ:mixture_model_supervised}) we wrote down the learner for the supervised case. For the online and batch case, the equivalent mixture models are:
\begin{align}
    Q^o(x^n) & = \int w(\theta) P_\theta(x^n) d\theta, \label{equ:mixture_model_online} \\
    Q^b(x_{n}\mymid x^{n-1}) & = \int_{\Theta} \underbrace{\frac{w(\theta) P_\theta(x^{n-1})}{\int w(\theta') P_{\theta'}(x^{n-1}) d\theta'}}_{w(\theta\mymid x^{n-1})} P_{\theta}(x_n \mymid x^{n-1}) d\theta, \label{equ:mixture_model_batch}
\end{align}
where $w(\theta)$ is a prior distribution on the weights that can be chosen freely, but we shall advocate for a uniform (or almost uniform) prior.

The choice of $Q^o(x^n)$ might seem puzzling at first --  we are working in the one-line setting and the sample $x^n$ is revealed to us only sequentially. However, $Q^o(x^n)$ as defined in (\ref{equ:mixture_model_online}) can be constructed sequentially by choosing the conditional distributions 
\begin{align}
Q^o(x_{t}\mymid x^{t-1}) 
& = \frac{Q^o(x^t)}{Q^o({x^{t-1}})} = \frac{\mathbbm{E}_{x_{t+1}, \cdots, x_n} \bigl[\int_{\theta} w(\theta) P_{\theta}(x^n) d\theta \bigr]}{\mathbbm{E}_{x_{t}, \cdots, x_n} \bigl[\int_{\theta'} w(\theta') P_{\theta'}(x^n) d\theta' \bigr]} 
= \frac{\int_{\theta} w(\theta) \bigl[\mathbbm{E}_{x_{t+1}, \cdots, x_n} \bigl[ P_{\theta}(x^n)\bigr] d\theta \bigr]}{\int_{\theta'} w(\theta') \bigl[\mathbbm{E}_{x_{t}, \cdots, x_n} \bigl[P_{\theta'}(x^n)\bigr] d\theta' \bigr]} \nonumber \\
& = \frac{\int_{\theta} w(\theta) P_{\theta}(x^t) d\theta}{\int_{\theta'} w(\theta') P_{\theta'}(x^{t-1}) d\theta'} 
 = \int \underbrace{\frac{w(\theta) P_\theta(x^{t-1})}{\int w(\theta') P_{\theta'}(x^{t-1}) d\theta'}  }_{w(\theta \mymid  x^{t-1})} P_\theta (x_{t}\mymid x^{t-1})  d \theta. \label{equ:local_mixture_model}
\end{align}
Here, $w(\theta \mymid x^{t-1})$ is the posterior distribution of $\theta$ at time $t$, given the observation $x^{t-1}$. Note in particular that $Q^o(x_{t}\mymid x^{t-1})$ is computable based on the knowledge of $x^{t-1}$ only.

Why are mixture models good learners? This is easiest to see in the realizable online setting, which we focus on here. A similar reasoning applies to the batch and supervised settings. 

Consider a probability assignment $\tilde{Q}(x^n)$ that is not a mixture. Let $Q$ be its projection onto the convex hull of $\Theta$, i.e., $Q$ is the distribution $\int_{\theta \in \Theta} w(\theta)P_\theta(x^n) d\theta$ that minimizes $D_{\text{KL}}(P_\theta\mydoublemid Q )$. The divergence projection theorem (see \citet{csiszar1975} and \citet[Theorem 11.6.1]{CoverThomas2012}) states that for all $P_\theta \in \Theta$:
\begin{align}
    D_{\text{KL}}(P_{\theta} \mydoublemid  \tilde{Q}) \geq D_{\text{KL}}(P_{\theta}\mydoublemid Q) + D_{\text{KL}}(Q \mydoublemid \tilde{Q}) \geq D_{\text{KL}}(P_{\theta}\mydoublemid Q ).
\end{align}
If the true model is $P_{\theta_0}$, recall that the regret, as a function of the learner, is defined as $\regret^o_r(\cdot,\theta)=\frac1n D_{\text{KL}}(P_{\theta_0} \mydoublemid \cdot)$. This shows that using a mixture model reduces the regret compared to a non-mixture assignment.

As a direct example, consider an online predictor that at each time $t$ picks the parameter $\theta^*_t$ that maximizes the likelihood $P_\theta(x^{t-1})$. This is the common empirical risk minimization (ERM) rule. Since this predictor is not a mixture, it can be improved.

Moreover, the exact weighting $w(\theta)$ used in the mixture is not very critical as long as it is spread over the entire parameter space. Let $Q(x^n)  = \int w(\theta) P_\theta(x^n) d\theta$ for some weight $w(\theta)$. Suppose instead we use another learner $\tilde{Q}$. By the lower bound of \citet[Theorem 2]{FederMerhavHierarchical}:
\begin{align}
\regret^o_r(\tilde{Q},\theta) \geq \regret^o_r(Q,\theta) - \frac{\gamma_n}{n}
\label{equ:strong_lower_bound}
\end{align}
for all $\theta \in \Theta \setminus B$, where $B$ is a subset with weight upper bounded by $w(B) \leq 2^{-\gamma_n}$.  With a suitable choice of $\gamma_n$ (which may increase with $n$ but stays negligible compared to $n \regret$), the measure of $B$ under $w(\theta)$ becomes insignificant (If $w(\theta)$ is uniform over the {\em allowable} parameters, $w(B)$ is the standard Lebesgue measure.) Thus, the performance of the mixture learner is essentially optimal for most $\theta$. We restate the proof for the convenience of the reader.

\vspace{0.25cm}
\begin{proof}
    \begin{align}
        1 =  \mathbbm{E}_{x^n \sim Q(x^n)} \Bigl[ \frac{\tilde{Q}(x^n)}{Q(x^n)} \Bigr] = & \int_{\Theta} w(\theta) \mathbbm{E}_{x^n \sim P_{\theta}(x^n)} \Bigl[ 2^{\left[ -\log Q(x^n) - (-\log \tilde{Q}(x^n) )  \right]} \Bigr] \nonumber \\ \stackrel{\text{Jensen and divergence definition}}\geq & \int_\Theta w(\theta) 2^{ \left[ D_{\text{KL}}(P_{\theta}(X^n)\mydoublemid Q(X^n)) - D_{\text{KL}}(P_{\theta}(X^n)\mydoublemid \tilde{Q}(X^n)) \right]}.
        \label{equ:exp_divergence}
    \end{align}
  Since $\regret^o_r(Q,\theta)=\frac1n D_{\text{KL}}(Q\mydoublemid P_\theta)$, the set $B$ is defined as
\begin{align*}    
B & =\left\{ \theta\in\Theta : D_{\text{KL}}(P_{\theta}(X^n)\mydoublemid Q(X^n)) - D_{\text{KL}}(P_{\theta}(X^n)\mydoublemid \tilde{Q}(X^n)) > \gamma_n \right \},
\end{align*}
or equivalently,
\begin{align}
B = \left\{ \theta\in\Theta \;:\; 2^{\left[ D_{\text{KL}}(P_{\theta}(X^n)\mydoublemid Q(X^n)) - D_{\text{KL}}(P_{\theta}(X^n)\mydoublemid \tilde{Q}(X^n)) \right]} > 2^{\gamma_n} \right \}  .      
\end{align}
By Markov's inequality and (\ref{equ:exp_divergence}), this gives $w(B)\leq {2^{-\gamma_n}}$. 
\end{proof}

\section{Regret Bounds -- Details}
\label{sec:regret_boundsdetails}
For the realizable online case, the following upper bound on the regret was already presented in \cite{Barron1987,Barron1988}. The motivation in those papers was to assess the consistency and convergence of Bayesian estimators of density functions. It was termed the ``Bayesian Index of Resolvability''. The following bounds are all derived for the agnostic case.

\subsection{Language Models -- Online Case}
\label{sec:regret_boundsonline}
For $\epsilon^2 \geq 0$, define the set $\Theta_0^o$ as:
\begin{align}
\Theta_0^o & = \left\{ \theta \in \Theta: \underbrace{\frac1n \mathbbm{E}_{x^n \sim P(x^n)} \Bigl[ \log \frac{P_{\theta_0}(x^n)}{P_{\theta}(x^n)} \Bigr]}_{=\frac1n D_{\text{KL}}(P_{\theta_0}(X^n) \mydoublemid P_{\theta}(X^n))\; \text{if realizable}} \leq \epsilon^2 \right\}.
\label{equ:Theta0_online} 
\end{align}
Let $w\left(\Theta_0^o\right)=\int_{ \Theta_0^o} w(\theta) d \theta$. We get
\begin{align*}
    \regret_a^o(Q,P) &   \stackrel{(\ref{equ:regret_online}) \& (\ref{equ:mixture_model_online})}{=}  \frac1n \mathbbm{E}_{x^n \sim P(x^n)} \Bigl[  \log \left( \frac{P_{\theta_0}(x^n)}{\int_{\Theta} w(\theta) P_{\theta}(x^n ) d\theta} \right)\Bigr]  \nonumber \\
    & \stackrel{\Theta_0^o \subseteq \Theta}  \leq \frac1n \mathbbm{E}_{x^n \sim P(x^n)} \Bigl[\log \left( \frac{P_{\theta_0}(x^{n})}{\int_{ \Theta_0^o} w(\theta) P_{\theta}(x^{n}) d \theta} \right) \Bigr]
    \nonumber \\
    &= 
    \frac1n \mathbbm{E}_{x^n \sim P(x^n)} \Bigl[\log \left( \frac{P_{\theta_0}(x^{n})}{\int_{\Theta_0^o} \tilde{w}(\theta)P_{\theta}(x^{n}) d\theta}  \frac{1}{\int_{ \Theta_0^o} w(\theta) d\theta} \right) \Bigr]
    \nonumber \\
    & 
    = \frac1n\mathbbm{E}_{x^n \sim P(x^n)} \Bigl[\log \left( \frac{P_{\theta_0}(x^{n})}{\int_{\theta \in \Theta_0^o} \tilde{w}_{\theta}(\theta) P_{\theta}(x^n) d \theta}\right)\Bigr] +\frac1n \mathbbm{E}_{x^n \sim P(x^n)} \Bigl[\log \frac{1}{w\left(\Theta_0^o\right)}  \Bigr]
    \nonumber \\
    & 
    \stackrel{\text{Jensen}}{\leq} \int_{\theta \in \Theta_0^o} \tilde{w}_{\theta}(\theta) \frac1n\mathbbm{E}_{x^n \sim P(x^n)} \Bigl[ \log \left( \frac{P_{\theta_0}(x^{n})}{ P_{\theta}(x^n) }\right)  \Bigr]d \theta +\frac1n \mathbbm{E}_{x^n \sim P(x^n)} \Bigl[\log \frac{1}{w\left(\Theta_0^o\right)}  \Bigr]
    \nonumber \\
   &
   \stackrel{(\ref{equ:Theta0_online})}{\leq}
       \epsilon^2 +  \frac1n \log \frac{1}{w\left(\Theta_0^o\right)} ,
\end{align*}
where we introduced the normalized measure $\tilde{w}_{\theta}(\theta) =  \frac{w(\theta)}{w\left(\Theta_0^o\right)} \mathbbm{1}_{\{\theta\in\Theta_0^o\}}$. Note that since the set $\Theta_0^o$ depends on $\epsilon^2$, so does $w(\Theta_0^o)$: As $\epsilon^2$ increases, $\log \frac{1}{w(\Theta_0^o)}$ can only decrease, and so there is an optimal value for the bound:
\begin{align}
\label{equ:online_general_bound_min_epsilon}    
\regret_a^o(Q,P) \leq \min_{\epsilon^2} \left[ \epsilon^2 +  \frac1n \log \frac{1}{w\left(\Theta_0^o\right)} \right].
\end{align}
This optimal value may depend on $n$.


\subsection{Language Models -- Batch Case}
\label{sec:regret_boundsbatch}

Similar to the online learning case, for $\epsilon^2\geq0$, define the set $\Theta_0^b(x^{n-1})$ as: 
\begin{align}
\Theta_0^b(x^{n-1}) & = \left\{ \theta \in \Theta: \underbrace{\mathbbm{E}_{x_n \sim P(x_n \mymid x^{n-1})} \Bigl[ \log \frac{P_{\theta_0}(x_n \mymid x^{n-1})}{P_{\theta}(x_n \mymid x^{n-1})} \Bigr]}_{=D_{\text{KL}}(P_{\theta_0}(X_n \mymid x^{n-1})\mydoublemid P_{\theta}(X_n \mymid x^{n-1}))\; \text{if realizable}}  \leq \epsilon^2 \right\}.
\label{equ:Theta0_batch} 
\end{align}
Note that this set depends on the training sequence $x^{n-1}$. Let $w\left(\Theta_0^b(x^{n-1})\mymid x^{n-1}\right)=\int_{\Theta_0^b(x^{n-1})} w(\theta\mymid x^{n-1} ) d\theta$. We get:
\begin{align}
    \regret_a^b(Q,P) &  \stackrel{(\ref{equ:regret_batch}) \& (\ref{equ:mixture_model_batch})}{=}  \mathbbm{E}_{x^n \sim P(x^n)} \Bigl[ \log \left( \frac{P_{\theta_0}(x_n\mymid x^{n-1})}{\int_{\Theta} w(\theta\mymid x^{n-1}) P_{\theta}(x_n \mymid x^{n-1}) d\theta} \right)  \Bigr] \nonumber \\
    & \stackrel{\Theta_0^b(x^{n-1}) \subseteq \Theta}  \leq \mathbbm{E}_{x^n \sim P(x^n)} \Bigl[ \log \left( \frac{P_{\theta_0}(x_{n}\mymid x^{n-1})}{\int_{\Theta_0^b(x^{n-1})} w(\theta\mymid x^{n-1})P_{\theta}(x_{n}\mymid x^{n-1})d \theta} \right) \Bigr]
    \nonumber \\
    &= 
    \mathbbm{E}_{x^n \sim P(x^n)} \Bigl[ \log \left( \frac{P_{\theta_0}(x_{n}\mymid x^{n-1})}{\int_{\Theta_0^b(x^{n-1})} \tilde{w}(\theta\mymid x^{n-1})P_{\theta}(x_{n}\mymid x^{n-1}) d\theta}  \frac{1}{ \int_{\Theta_0^b(x^{n-1})} w(\theta\mymid x^{n-1} ) d\theta}\right) \Bigr]
    \nonumber \\
    & 
    = \mathbbm{E}_{x^n \sim P(x^n)} \Bigl[\log \left( \frac{P_{\theta_0}(x_{n}\mymid x^{n-1})}{\int_{\Theta_0^b(x^{n-1})} \tilde{w}(\theta\mymid x^{n-1})P_{\theta}(x_{n}\mymid x^{n-1}) d \theta}\right) \Bigr]+ \nonumber \\
    & + \mathbbm{E}_{x^{n-1} \sim P(x^{n-1})} \Bigl[ \log \frac{1}{w\left(\Theta_0^b(x^{n-1})\mymid x^{n-1}\right)} \Bigr]
    \nonumber \\
    &
    \stackrel{\text{Jensen}}{\leq} \mathbbm{E}_{x^{n-1} \sim P(x^{n-1})} \Bigl[\int_{\Theta_0^b(x^{n-1})} \tilde{w}(\theta\mymid x^{n-1}) \mathbbm{E}_{x_n \sim P(x_n \mymid x^{n-1})} \Bigl[ \log \left( \frac{P_{\theta_0}(x_{n}\mymid x^{n-1})}{P_{\theta}(x_{n}\mymid x^{n-1}) }\right) \Bigr] d \theta \Bigr] + \nonumber \\
    & + \mathbbm{E}_{x^{n-1} \sim P(x^{n-1})} \Bigl[\log \frac{1}{w\left(\Theta_0^b(x^{n-1})\mymid x^{n-1}\right)} \Bigr] \label{equ:batch_general_no_epsilon}
 \\
    &
    \stackrel{(\ref{equ:Theta0_batch})}{\leq} 
      \epsilon^2 + \mathbbm{E}_{x^{n-1} \sim P(x^{n-1})} \Bigl[ \log \frac{1}{w\left(\Theta_0^b\mymid x^{n-1}\right)} \Bigr], \label{equ:batch_general_epsilon}
\end{align}
where we introduced the normalized measure 
$\tilde{w}(\theta \mymid x^{n-1}) = 
   \frac{w(\theta \mymid  x^{n-1})}{\int_{\Theta_0^b}w(\theta \mymid x^{n-1}) d\theta} \mathbbm{1}_{\{\theta\in\Theta_0^b (x^{n-1}\}}$. 
The bound we got can be written as follows:
\begin{align}
\begin{split}
    \regret_a^b(Q,P) &\leq \mathbbm{E} \left[ \int \tilde{w}(\theta \mymid x^{n-1} \log \left( \frac{P_{\theta_0}(x_n \mymid x^{n-1}}{P_{\theta}(x_n \mymid x^{n-1}} \right) \right] - \mathbbm{E}_{x^{n-1} \sim P(x^{n-1})} \Bigl[ \log \left(w(\Theta_0^b \mymid x^{n-1}) \right) \Bigr] 
    \\
    &\leq 
    \epsilon^2 - \mathbbm{E}_{x^{n-1} \sim P(x^{n-1})} \Bigl[ \log \left(w(\Theta_0^b \mymid x^{n-1}) \right)  \Bigr].
    \end{split}
\end{align}

\subsection{Supervised Learning}
\label{sec:regret_bound_supervised}
Similar to the previous two cases, for $\epsilon^2 \geq 0$, define the set $\Theta_0^s$ as: 
\begin{align}
\Theta_0^s & = \left\{ \theta \in \Theta: \underbrace{\mathbbm{E}_{(x, y) \sim P_X(x) P(y \mymid x)} \Bigl[ \log \left( \frac{P_{\theta_0}(y \mymid  x)}{P_{\theta}(y \mymid x) }\right) \Bigr]}_{=D_{\text{KL}}(P_{\theta_0}(x \mymid y) P_X(x)\mydoublemid P_{\theta}(x \mymid y) P_X(x))\; \text{if realizable}}  \leq \epsilon^2 \right\}. 
\label{equ:Theta0_supervised}
\end{align}
Let $w\left(\Theta_0^s\mymid {\mathcal S}\right)=\int_{\Theta_0^s} w(\theta\mymid {\mathcal S} ) d\theta$. Let ${\mathcal S}=\{(x_i, y_i)\}_{i=1}^{n} \sim \prod_{i=1}^{n} P_X(x_i) P(y_i \mid x_i) $. We get
\begin{align*}
\regret^s_a(Q,P) & \stackrel{(\ref{equ:regret_supervised_agnostic}) \& (\ref{equ:mixture_model_supervised})}{=} 
     \mathbbm{E}_{{\mathcal S}, (x, y) \sim  P_X(x) P(y \mymid x)} \Bigl[
     \log \left( \frac{P_{\theta_0}(y\mymid x)}{\int_{\Theta} w(\theta \mymid {\mathcal S}) P_{\theta}(y \mymid x) d\theta)} \right) \Bigr] \nonumber \\
    & \stackrel{\Theta_0^s \subseteq \Theta} \leq \mathbbm{E}_{{\mathcal S}, (x, y) \sim  P_X(x) P(y \mymid x)} \Bigl[ \log \left( \frac{P_{\theta_0}(y \mymid x)}{\int_{\Theta_0^s} w(\theta \mymid {\mathcal S}) P_{\theta}(y \mymid x) d\theta)} \right) \Bigr]  \nonumber \\
    \nonumber \\
    & = \mathbbm{E}_{{\mathcal S}, (x, y) \sim  P_X(x) P(y \mymid x)} \Bigl[ \log \left( \frac{P_{\theta_0}(y \mymid x)}{\int_{\Theta_0^s} \tilde{w}(\theta \mymid {\mathcal S}) P_{\theta}(y \mymid x) d\theta} \frac{1}{\int_{\Theta_0^s} w(\theta \mymid {\mathcal S}) d\theta} \right) \Bigr] \nonumber \\
    \nonumber \\
    & 
    = \mathbbm{E}_{{\mathcal S}} \Bigl[  \mathbbm{E}_{(x, y) \sim  P_X(x) P(y \mymid x)} \Bigl[ \log \left( \frac{P_{\theta_0}(y \mymid  x)}{\int_{ \Theta_0^s} \tilde{w}(\theta\mymid {\mathcal S})P_{\theta}(y \mymid x) d \theta}\right) \Bigr] \Bigr]+
    \nonumber \\
    & \phantom{=\;} + \mathbbm{E}_{{\mathcal S}} \Bigl[  \log \frac{1}{w\left(\Theta_0^s\mymid {\mathcal S}\right)} \Bigr]
    \\
    &
        \stackrel{\text{Jensen}}{\leq} \mathbbm{E}_{{\mathcal S}} \Bigl[ \int_{ \Theta_0^s} \tilde{w}(\theta\mymid {\mathcal S}) \mathbbm{E}_{(x, y) \sim  P_X(x) P(y \mymid x)} \Bigl[ \log \left( \frac{P_{\theta_0}(y \mymid  x)}{P_{\theta}(y \mymid x) }\right) \Bigr] d \theta  \Bigr]+
    \nonumber \\
    & \phantom{=\;} + \mathbbm{E}_{{\mathcal S}} \Bigl[\log \frac{1}{w\left(\Theta_0^s\mymid {\mathcal S}\right)} \Bigr]
    \nonumber \\
    &
    \stackrel{ (\ref{equ:Theta0_supervised})}{\leq} 
    \epsilon^2 + \mathbbm{E}_{{\mathcal S}} \Bigl[\log \frac{1}{w\left(\Theta_0^s\mymid {\mathcal S}\right)} \Bigr] , 
\end{align*}
where we introduced the normalized measure $\tilde{w}(\theta\mymid {\mathcal S}) =  \frac{w(\theta\mymid {\mathcal S})}{w\left(\Theta_0^s \mymid {\mathcal S}\right)} \mathbbm{1}_{\{\theta\in\Theta_0^s\}}$. We thus arrive at the following bounds:
\begin{align} 
    \regret^s_a(Q,P) & \leq \mathbb{E}_{\mathcal S} \left[ \int_{\theta}\tilde{w(\theta \mymid \mathcal{S})} \log \left( \frac{P_{\theta_0}(y \mymid x)}{P_{\theta}(y \mymid x)} \right) \right] - \mathbb{E}_{\mathcal S}\left[\log(w(\Theta_0 \mymid \mathcal{S})\right] \label{equ:supervised_general_bound_no_epsilon} \\
    & \leq \epsilon^2 - \mathbb{E}_{\mathcal S}\left[\log(w(\Theta_0 \mymid \mathcal{S})\right] \label{equ:supervised_general_bound_epsilon}
\end{align}
Since these bounds are valid for every $\epsilon^2$, we can minimize over $\epsilon^2$ to attain the tightest possible bound.

\section{Spectrum-Based Regret Bounds -- Details for Realizable Case}
\label{sec:auxiliary_complexity_measures_details_realizeable}
In this appendix, we focus on the {\em realizable} case, where the data-generating distribution 
$P(\cdot)$ lies within the model class $\Theta$, i.e., $P(\cdot) = P_{\theta_0}(\cdot)$ for some $\theta_0$. Hence, as defined in a  few places above, we will write  $\regret(Q,\theta_0)$ instead of the more general $\regret(Q,P)$. Our analysis covers online learning, batch learning, and supervised learning. The agnostic case will be treated in Appendix~\ref{sec:auxiliary_complexity_measures_details_agnostic}.

\subsection{Language Models -- Online Case}
\label{sec:auxiliary_complexity_measures_online_details_realizeable}
The basic idea of the spectrum-based bounds is as follows. In general, the full structure of $\Theta_0^o$---which may be required for evaluating the regret bound---can 
be intricate, potentially consisting of disjoint regions in parameter space due to model symmetries, 
a feature commonly observed in deep neural networks (DNNs). 
Nevertheless, we can obtain a valid upper bound on the regret by analyzing the {\em local} behavior of 
$D_{\text{KL}}\!\left(P_{\theta_0}\,\Vert\,P_\theta\right)$ for $\theta$ in a neighborhood of $\theta_0$, 
as characterized by the Fisher information matrix (FIM):
\begin{align}
\label{equ:fisher_information_matrix}
I_n(\theta) 
   &= \mathbbm{E}_{\theta}\!\left[(\nabla_{\theta} \log P_{\theta}(x^n))(\nabla_{\theta} \log P_{\theta}(x^n))^\top\right].
\end{align}
This characterization follows from the local expansion
\begin{align}
 D_{\mathrm{KL}}(P_{\theta_0}\,\Vert\,P_\theta) 
   &= (\theta-\theta_0)^\top \frac{I_n(\theta_0)}{2} (\theta-\theta_0) 
   + O\!\left(\|\theta-\theta_0\|^3\right),
 \label{equ:expansion}
\end{align}
which holds under standard regularity conditions. Note that $I_n(\theta_0)$ is Hermitian, hence it admits an orthonormal eigenbasis and real eigenvalues. Further, $I_n(\theta_0)$ is positive semidefinite, i.e., the eigenvalues are in fact non-negative. Thus, the local behavior of $D_{\text{KL}}$ is determined by the spectrum of the FIM. First, we will prove this of the online case. The same basic idea will later be applied also to the batch case and the supervised case.

\begin{theorem}
\label{the:online_local_properties}
Suppose $\Theta$ is contained within a $d$-dimensional ball of radius $R$, and the prior $w(\theta)$ is uniform over $\Theta$. Further assume that the eigenvalues of the Fisher information matrix $I_n(\theta)$, denoted $\lambda_1 \geq \lambda_2 \geq \cdots \geq \lambda_d \geq 0$, satisfy
\[
\lambda_{k+1} \leq \frac{k}{R^2 \log(e)} \quad \text{for some } k.
\]
Then
\begin{align}
    \regret_r^o(Q,\theta_0) 
    &\leq \frac{1}{2n}\biggl[\frac{k}{\log(e)} 
        + k \log\!\left(\frac{\log(e)}{k}\right) 
        + \sum_{i=1}^k \log\!\bigl(R^2 \lambda_i\bigr)\biggr].
    \label{equ:online_regret_general_rad}
\end{align}
\end{theorem}

\begin{proof}
To characterize $\Theta_0$ defined in~\eqref{equ:Theta0_online}, we use the local expansion~\eqref{equ:expansion}.  
Let $d$ denote the dimensionality of $\Theta$, and let $\{\lambda_i\}_{i=1}^d$ be the ordered eigenvalues of $I_n(\theta_0)$.  
From~\eqref{equ:expansion}, ignoring higher-order terms, the set $\Theta_0$ is approximately an ellipsoid with principal axis lengths
\[
\left\{\frac{\sqrt{2n}\,\epsilon_n}{\sqrt{\lambda_i}}\right\}_{i=1}^{d},
\]
where $\epsilon_n$ captures the dependence on $n$. Hence the volume of $\Theta_0$, denoted $V(\Theta_0)$, is approximately
\begin{align}
    V(\Theta_0) 
        \;\cong\; \frac{(\sqrt{2n}\,\epsilon_n)^d}{\prod_{i=1}^{d} \sqrt{\lambda_i}} 
        \, V\!\left(S(1)\right),
\end{align}
where $S(1)$ is the unit $d$-dimensional ball and $V(\cdot)$ denotes volume.

Assume now that the prior is uniform over a $d$-dimensional volume contained in a ball of radius $R$ centered at $\theta_0$. Then
\[
w(\Theta_0) \;\geq\; \frac{V(\Theta_0 \cap S(R))}{V(S(R))}.
\]
If along some axis the ellipsoid radius exceeds $R$, the effective radius is capped at $R$. Thus the effective axes are
\[
\left\{\min\!\left\{\frac{\sqrt{2n}\,\epsilon_n}{\sqrt{\lambda_i}},\,R\right\}\right\}_{i=1}^d.
\]
Let $k$ be the largest index such that $\frac{\sqrt{2n}\,\epsilon_n}{\sqrt{\lambda_k}} \leq R$. We call $k=k(n)$ the {\em effective number of parameters}. It follows that
\begin{align}
\label{equ:ratio}
w(\Theta_0) 
   &= \frac{V(\Theta_0 \cap S(R))}{V(S(R))} \\
   &\cong \frac{(\sqrt{2n}\,\epsilon_n)^k \, R^{d-k}}{\prod_{i=1}^k \sqrt{\lambda_i}} 
      \cdot \frac{V(S(1))}{R^d V(S(1))} \nonumber \\
   &= \frac{(\sqrt{2n}\,\epsilon_n)^k}{R^k \prod_{i=1}^k \sqrt{\lambda_i}}. \nonumber
\end{align}
Substituting~\eqref{equ:ratio} into~\eqref{equ:online_general_bound_min_epsilon} gives
\begin{align}
\regret^o_r \;\leq\; 
    \min_{\epsilon_n} \Biggl[ 
        \epsilon_n^2 
        - \frac{k}{2n}\log(2n \epsilon_n^2) 
        + \frac{1}{2n}\sum_{i=1}^k \log(R^2 \lambda_i)
    \Biggr].
\end{align}
Differentiating with respect to $\epsilon_n^2$ shows the minimizer is
\[
\epsilon_n^2 \;=\; \frac{k}{2n\log(e)}.
\]
Substituting this back yields
\begin{align*}
\regret 
   &\leq \frac{1}{2n}\biggl[\frac{k}{\log(e)} 
        + k \log\!\left(\frac{\log(e)}{k}\right) 
        + \sum_{i=1}^k \log(R^2 \lambda_i)\biggr],
\end{align*}
as claimed.
\end{proof}
To further interpret the result, assume that for all $i \in \{1,\dots,d\}$ we have $\theta_i \in [-a,a]$.  
Since the ball of radius $R$ must enclose the cube $[-a,a]^d$, we obtain a lower bound $R^2 \geq a^2 d$.  
In this case,
\begin{align*}
\regret^o_r \;\leq\; \frac{1}{2n}\Bigl[\frac{k}{\log(e)} 
   + k \log \bigl(d a^2 \tfrac{\log(e)}{k}\bigr) 
   + \sum_{i=1}^k \log(\lambda_i^{(n)})\Bigr].
\end{align*}
Since the FIM is defined on an $n$-tuple $X^n$, its eigenvalues typically scale linearly with $n$. Writing $\lambda_i=\lambda_i^{(n)}$, we expect $\lambda_i^{(n)} \approx n \tilde{\lambda}_i$ for constants $\tilde{\lambda}_i$. Thus the dominant contribution in~\eqref{equ:online_regret_general_rad} is
\begin{align}
\frac{1}{2n}\bigl[k \log(d) + k \log(n)\bigr]. 
\label{equ:dominant}
\end{align}

If the model has $k$ free parameters, classical information-theoretic results (see \citet{Rissanen1984}) show that the smallest achievable regret is $\tfrac{k \log(n)}{2n}$, which matches the second term in~\eqref{equ:dominant}. Therefore $k$ can be interpreted as the number of \emph{effective} parameters: the term $\tfrac{k \log(n)}{2n}$ corresponds to the number of bits required to specify the parameter values. Since $\epsilon^2 = O(1/n)$, or $\epsilon = O(1/\sqrt{n})$, each parameter (assumed to lie in a finite range) can be quantized into $O(\sqrt{n})$ levels, requiring about $\tfrac12 \log n$ bits per parameter. Consequently, the regret naturally admits an interpretation as the \emph{code length} for describing parameter values, given the identification of these $k$ effective parameters.

Continuing this coding analogy, the first term in~\eqref{equ:dominant} corresponds to the code length required to specify which $k$ of the $d$ available parameters are active. This requires
\[
\log_2\!\binom{d}{k} 
   = \log_2 \left( \frac{d!}{k!\,(d-k)!} \right)
\]
bits. When $k \ll d$, $\binom{d}{k} \approx d^k$, so the description length is approximately $k \log_2(d)$ bits, which is consistent, up to a factor $\tfrac12$, with the first term in~\eqref{equ:dominant}.

This comparison is heuristic rather than exact. For example, one must also encode the value of $k$ itself, which can be done using Elias’s universal code at an additional cost between $\log k$ and $2\log k$ bits (see \citet{1055349}), corresponding to the extra terms in~\eqref{equ:online_regret_general_rad}.

Finally, note that \citet{ClarkeBarron1994} also analyzed the local regret of mixture distributions with general priors $w(\theta)$ using Laplace approximation. Their focus was on finding the least favorable prior, thereby equalizing regret across $\theta$. As argued here, regret need not be equalized. Substituting a uniform prior into their framework yields asymptotic regret expressions consistent with ours, though their analysis is asymptotic and restricted to the online case.

\subsection{Supervised Case: Proof of Theorem \ref{the:supervised_local_properties_theorem}}
\label{sec:supervised_proof}
\begin{proof}
First, set $\epsilon^2$ sufficiently large so that $\Theta_0 = \Theta$, and thus $\mathbbm{E}_{{\mathcal S}} \Bigl[\log \left( {w\left(\Theta_0^s\mymid {\mathcal S}\right)} \right))\Bigr] = 0$, and $\tilde{w}(\theta \mymid \mathcal{S}) = w(\theta \mymid \mathcal{S})$. The bound  (\ref{equ:supervised_general_bound_no_epsilon}) then simplifies to
\begin{align}
    \regret_r^s(Q,\theta_0) \leq \mathbb{E}_{{\mathcal S}, x} \left[\int_{\Theta} w(\theta \mymid \mathcal{S})D \left( P_{\theta_0}(y \mymid x) \mydoublemid P_{\theta}(y \mymid x) \right) d\theta \right]
    \label{equ:supervised_whole_Theta_average_kl},
\end{align}
where in the averaging, both the training data inputs and the test data inputs are distributed according to $P_{X}$, and the training outcomes are distributed according to $P_{\theta_0}(y \mymid x)$.
The KL-divergence in the vicinity of $\theta_0$ can be approximated using $I(\theta_0)$, the FIM at $\theta_0$:
\begin{align}
    \mathbb{E}_{x \sim P_{X}} \Bigl[D_{\text{KL}}(P_{\theta_0}(y \mymid x) \mydoublemid P_{\theta}(y \mymid x)) \Bigr] = \frac{(\theta_0 - \theta)^T I(\theta_0) (\theta_0 - \theta)}{2} + O(\|\theta_0 - \theta\|^3), \label{equ:supervised_expansion}
\end{align}
where now $I(\theta_0)$ is the average FIM with respect to $x$.
Assume that the eigenvalues of $I(\theta_0)$ are sorted in decreasing order, $\lambda_1 \geq \lambda_2 \geq ... \geq \lambda_d$. In addition, assume that $\Theta$ is a $d$-dimensional ball of radius $R$, and that there exists some $\alpha > 0$ and $k \in \naturals$, $k \ll \min\{n, d\}$, so that
\begin{align}
\lambda_{k+1} \leq \frac{\alpha}{2R^2}. \label{equ:supervised_condition}
\end{align} 
If we denote by $\tilde{\theta}[i]$ the projection of $\theta$ on the $i$-th eigenvector we get:
\begin{align}
    \frac{(\theta_0 - \theta)^T I(\theta_0) (\theta_0 - \theta)}{2} = \frac{\sum_{i=1}^d \lambda_i(\tilde{\theta_0}[i] - \tilde{\theta}[i])^2} {2} =  \frac{\sum_{i=1}^k \lambda_i(\tilde{\theta_0}[i] - \tilde{\theta}[i])^2} {2} + \delta, \label{equ:supervised_I_bound}
\end{align}
where $\delta = \frac{\sum_{i=k+1}^d \lambda_i(\tilde{\theta_0}[i] - \tilde{\theta}[i])^2} {2} \leq \alpha$ because $||\theta-\theta_0||^2 \leq 2R^2$.
Next, we invoke Laplace's approximation, which states that 
\begin{align}
w(\theta \mymid \mathcal{S}) \sim \mathcal{N}\left(\hat{\theta}(\mathcal{S}), \frac{I^{-1}(\hat{\theta}(\mathcal{S}))}{n} \right), \label{equ:supervised_laplace}
\end{align} where $\hat{\theta}(\mathcal{S})$ is the maximum likelihood estimator of $\theta$ given $\mathcal{S} = \{(x, y)\}_{t=1}^{n-1}$. If we further assume that $I(\theta_0)$ is smooth around $\theta_0$ so that $I_{\hat{\theta}(\mathcal{S})} \approx I(\theta_0)$, we get:

\begin{align}
     \regret_r^s(Q,\theta_0) - \delta &\stackrel{(\ref{equ:supervised_whole_Theta_average_kl}) \& (\ref{equ:supervised_expansion}) \& (\ref{equ:supervised_I_bound})}{\leq} \mathbb{E}_{\mathcal{S}} \left[ \int_{\Theta} w(\theta\ \mathcal{S})\frac{\sum_{i=1}^k \lambda_i(\tilde{\theta_0}[i] - \tilde{\theta}[i])^2} {2}\right] 
     \nonumber \\
     &\stackrel{(\ref{equ:supervised_laplace})}{\approx} 
     \mathbb{E}_{\mathcal{S}} \left[ \mathbb{E}_{\theta \sim \mathcal{N}\left(\hat{\theta}(\mathcal{S}), \frac{I^{-1}(\hat{\theta}(\mathcal{S}))}{n-1} \right)} \left[ \frac{\sum_{i=1}^k \lambda_i(\tilde{\theta_0}[i] - \tilde{\theta}[i])^2} {2} \right] \right] 
    \nonumber \\
    &=
  \mathbb{E}_{\mathcal{S}} \left[ \mathbb{E}_{\theta \sim \mathcal{N}\left(\hat{\theta}(\mathcal{S}), I^{-1}(\hat{\theta}(\mathcal{S})) \right)} \left[ \frac{\sum_{i=1}^k \lambda_i(\tilde{\theta_0}[i] - \tilde{\theta}[i])^2} {2n} \right] \right]
    \nonumber \\
    &\approx
    \mathbb{E}_{\mathcal{S}} \left[  \mathbb{E}_{\theta \sim \mathcal{N}\left(\hat{\theta}(\mathcal{S}), I^{-1}_{\theta_0} \right)} \left[ \frac{\sum_{i=1}^k \lambda_i(\tilde{\theta_0}[i] - \tilde{\theta}[i])^2} {n} \right] \right]
    \nonumber \\
    &=
    \mathbb{E}_{\mathcal{S}} \left[ \left(\frac{k + \sum_{i=1}^k (\tilde{\theta_0}[i] - \tilde{\hat{\theta}}(\mathcal{S})[i])^2  }{2n} \right) \right] =
    \frac{k}{2n} + o(\frac{1}{n}),
\end{align}
where, again, in the last equality we used $\hat{\theta} \sim \mathcal{N}\left(\theta_0, \frac{I(\theta_0)}{n}\right)$, and in the one before we used the average of a non-centered $\chi^2$ distribution.
\end{proof}

Recall that in this setting, we assumed that the inputs follow some distribution $P_X$. Fortunately, the learner $Q(y \mymid x, \mathcal{S})$ itself does not depend on the assumed $P_{X}$. But the performance  $\mathbbm{E}_{X \sim P_X} \left[I(\theta_0) \right]$ does. It will be interesting to obtain extensions to the above result that are based on the samples $\mathcal{S}$ rather than the actual distribution $P_{X}$.

It should be noted that when $\alpha \ll \frac{k}{2n}$, the regret is bounded by $\frac{k}{2n} + o \left( \frac{1}{n} \right)$. In this case, $k$ can be interpreted as the effective dimension since in the general case, where the eigenvalues are not so small, the regret is of order $\frac{k}{2n}$.

Note that the derivation here is rather different than the online equivalent described in \ref{sec:auxiliary_complexity_measures_online_details_realizeable}, since here we are essentially analyzing the expected KL-divergence over the entire parameter space $\Theta$, which decays as $\frac{k}{2n}$ since the posterior in the effective dimensions of $k$ converges to $\theta_0$. In the online case, such an analysis over $\Theta$ would yield a very large average divergence. On the other hand, we could have followed for the batch case a similar derivation as for the online case, and furthermore, look for the optimal $\epsilon^2$. Since we have reached what is probably the optimal bound of $\frac{k}{2n}$ for our choice of large $\epsilon^2$, following the online derivation and finding the optimal $\epsilon^2$ will not provide a better bound.

Interestingly, one can choose other values for $\epsilon^2$ that will essentially lead to the same result: First, note that for any $\epsilon^2, \mathcal{S}$, we have:
\begin{align}
     \int_{\Theta^s_0} \tilde{w}(\theta \mymid \mathcal{S}) \mathbb{E}_{x} \left[D \left( P_{\theta_0}(\cdot \mymid x) \mydoublemid {P_{\theta}(\cdot \mymid x)} \right) \right] \leq \int_{\Theta} w(\theta\mymid \mathcal{S}) \mathbb{E} _x \left[ D \left( P_{\theta_0}(\cdot \mymid x) \mydoublemid {P_{\theta}(\cdot \mymid x)} \right) \right]
\end{align}
Now, assuming that Laplace's approximation holds, and that $I(\theta_0) \approx I(\hat{\theta})$, we have:

\begin{align}
    w(\theta \mymid\mathcal{S}) \sim \mathcal{N}\left(\hat{\theta}, \frac{I^{-1}(\theta_0)}{n-1}\right)
\end{align}
On the other hand, utilizing the definition of $\Theta_0^s$ and \eqref{equ:supervised_I_bound}, we have:
\begin{align}
    \Theta_0^b \subset \{ \theta: \frac{\sum_{i=1}^k \lambda_i(\tilde{\theta_0}[i] - \tilde{\theta}[i])^2} {2} \leq \epsilon^2 - \delta \}.
\end{align}
Thus, we get that $w(\Theta^b_0 \mymid \mathcal{S})$ is the cumulative distribution function of a non-center $\chi^2$ distribution with $k$ degrees of freedom evaluated at $\epsilon^2 n$, where the $i$-th coordinate of the center is $\tilde{\theta_0} - \tilde{\hat{\theta}}$.

If we now choose, for example, $\epsilon^2 > \frac{\alpha k \log(n)}{n}$ for some $\alpha > 1$, we get that $\epsilon^2 n = k \alpha \log(n)$. Since $\hat{\theta} \sim \mathcal{N}\left(\tilde{\theta_0}, \frac{I^{-1}(\theta_0)}{n-1}\right)$, the contribution of the averages to the $\chi^2$ distribution is negligible, and we can estimate a lower bound on $w(\Theta_0^s \mymid \mathcal{S})$ as the probability that a $\chi^2$ distribution with $k$ degrees of freedom is larger than $k \alpha \log(n)$, which by \cite{laurent2000adaptive} is bounded by
\begin{align}
    w(\Theta_0^s \mymid \mathcal{S}) \geq  1 - e^{-\alpha \log(n)} = 1 - \frac{1}{n^\alpha}
\end{align}
given that $\alpha k \log(n)  > 2\sqrt{k\alpha\log(n)} + 2\alpha \log(n)$. Thus, 
\[\mathbb{E}_{x^{n-1}} \left[-\log(w(\Theta_0^b \mymid x^{n-1}) \right ] \leq -\log(1-\frac{1}{n^\alpha}) = o(\frac{1}{n})\]
and we get again a bound over the regret which behaves as $\frac{k}{2n} + o(\frac{1}{n})$.

These $\epsilon^2$, as well as the large $\epsilon^2$ considered above, may not be the optimal value that minimizes the upper bound. Nevertheless, they attain the correct main term $k/2n$ of the bound. 

\subsection{Language Models -- Batch Learning}
\label{sec:auxiliary_complexity_measures_batch_details_realizeable}

In this section, we shall discuss possible ways to derive bounds over the regret for the batch learning problem based on local properties of the hypothesis class around the true data-generating model $\theta_0$. 

One possible way to derive a spectrum-based regret bound for the batch learning case might be to utilize the results for the online case. The online case and the batch case are related in the following simple manner:
\begin{align}
    \regret_r^b(Q,\theta_0;n) = n\regret_r^o(Q,\theta_0;n) - (n-1)\regret_r^o(Q,\theta_0;n-1), 
    \label{equ:batch_versus_online}
\end{align}
where $\regret_r^o(Q,\theta_0;n)$ is given by (\ref{equ:regret_online}) and $\regret_r^b(Q,\theta_0;n)$ is defined in (\ref{equ:regret_batch}).\footnote{We added the length $n$ explicitly to the original expressions in (\ref{equ:regret_online}) and  (\ref{equ:regret_batch}) since in the current context $n$ plays a crucial role.} 
If we were to blindly plug the upper bound of the online case given in (\ref{equ:online_regret_general_rad}) into the two terms on the right of (\ref{equ:batch_versus_online}) we would get that $\regret_r^b(Q,\theta_0;n) \sim \frac12 \sum_{i=1}^{k} (\log(\lambda_i^{(n)})-\log(\lambda_i^{(n-1)})) \sim
\frac12 k (\log(n)-\log(n-1)) \sim  \frac12 \frac{k}{n}$, where we assumed, as explained in the previous section, that $\lambda_i^{(n)} \approx n \tilde{\lambda}_i$.
Of course, mathematically, this is not a valid derivation since we took the difference of two upper bounds, but the result gives us what we expect to be the ``correct'' bound. 

Rewrite (\ref{equ:batch_versus_online}) as $\regret_r^o(Q,\theta_0;n)  = \frac1n \sum_{i=1}^{n} \regret_r^b(Q,\theta_0;i)$.
From this, we conclude that the upper bound for the online case in (\ref{equ:online_regret_general_rad}) is an upper bound on the \emph{average} of the batch case, where the average is with respect to $n$. The dominant term of this average bound is $\frac1{2n}[k \log(d)+k \log(n)]$. This bound is worse by a factor $\log(d n)$ than what we expect from analogies to universal prediction. 

Without an assumption on the underlying true hypothesis $P_{\theta_0}(x^n)$ we cannot expect a point-wise bound as the following simple examples shows. Assume that for the underlying distribution, the first $n/2$ samples are independent of the remaining $n/2$ samples. Then at time $n/2+1$ we cannot expect a good bound. However, in ``real'' systems it is reasonable to assume that the underlying distribution fulfills a stationarity constraint so that over time predictions only become better.

\subsubsection*{Batch Learning: Memoryless Hypothesis Classes}

Consider first a simplified case of batch learning where the hypothesis class is memoryless, i.e., $\forall{\theta, x^n}: P_{\theta}(x^n) = \prod_{t=1}^n P_{\theta}(x_t)$. In this case, it is rather straightforward to note that Theorem \ref{the:supervised_local_properties_theorem} can be applied to get a valid upper bound on the regret. Another way to bound the batch learning regret for memoryless hypothesis classes is to utilize the online to batch conversion proposed and analyzed in Appendix \ref{sec:supervised_online_to_batch}, which perhaps attains a looser upper bound but requires fewer assumptions.

\subsubsection*{Batch Learning: The Case with Memory}
\label{sec:with_memory_discussion}
We now turn our attention to the more complicated scenario where the hypothesis class consists of probability assignments that might depend on the past $m$ outcomes, i.e. $P_{\theta}(x_n \mymid x^{n-1}) = P_{\theta}(x_n \mymid x^{n-1}_{n-(m+1)})$, where $x^{n-1}_{n-(m+1)} = x_{n - m - 1},  x_{n-m}, ..., x_{n-1}$. In this case, one cannot simply use the derivation in Theorem \ref{the:supervised_local_properties_theorem}. The reason is that now, the final $m$ samples influence both $w(\theta \mymid x^{n-1})$ and the relevant divergence:
\begin{align}
    D \left( P_{\theta_0}(x_n \mymid x^{n-1}_{n-(m+1)}) \mydoublemid P_{\theta}(x_n \mymid x^{n-1}_{n-(m+1)}) \right) = \frac{(\theta_0 - \theta)^T I(\theta_0 \mymid x^{n-1}_{n-(m+1)}) (\theta_0 - \theta)}{2} + O(\|\theta_0 - \theta\|^3) 
\end{align}
where now $I(\theta \mymid x^{n-1}_{n-(m+1)})$ is the FIM at $\theta_0$ given $x^{n-1}_{n-(m+1)}$. 

Nevertheless, there are two settings in which one can follow the above derivation that essentially capture most of the relevant applications: First, if we consider large language models, then the context on which $x_n$ is predicted is usually not included in the training set. In this case, we might denote the context by $\tilde{x^m}$, which will be independent from the training set $x^{n-1}$ used to estimate the MLE $\hat{\theta}(x^{n-1})$.

Another setting where the derivation in Theorem \ref{the:supervised_local_properties_theorem} holds although there is dependency upon $x^{n-1}_{n-(m+1)}$ is when $m \ll n$, and that the probabilities in $\theta_0$ are such that after some $k \ll n$ samples, the dependency between the symbols prior to any sequence of length $k$ is independent from the symbols that follows those $k$ symbols. In this case, if indeed $k \ll n$, then the effect of the last $k$ samples on the $w(\theta \mymid (x^{n-1})$ is negligible, and we can again apply the derivation leading to the bound in Theorem \ref{the:supervised_local_properties_theorem}. 

\subsection{Batch and Supervised Learning: Multi-modal Case}

One should note that Theorem \ref{the:supervised_local_properties_theorem} analyzes the regret using local properties of the KL-divergence around $\theta_0$, essentially assuming that the posterior over $\theta$, either $w(\theta \mymid x^{n-1})$ or $w(\theta \mymid \mathcal{S})$, is concentrated around $\theta_0$. One might wonder if this is a valid assumption in a general, over-parameterized case, where there could be many different values $\theta$'s, which we will denote by $\hat{\theta}_1, \ldots, \hat{\theta}_k$, attaining either the maximum likelihood probability over the training set, $x^{n-1}$, or $\mathcal{S}$, which should also be taken into account. In this case, it turns out that a better approximation for the posterior is a Gaussian mixture:

\begin{align}    
w(\theta \mymid x^{n-1}) = \sum_{i=0}^{k} w_i \mathcal{N}\left(\hat{\theta}_i, I_{\hat{\theta}_i} \right)
\end{align}

where $\hat{\theta}_0$ is assumed to be close to $\theta_0$ while $\hat{\theta}_1 \ldots \hat{\theta}_k$ might be far from $\theta_0$. Naturally, a similar approximation can be considered for $w(\theta \mymid \mathcal{S})$ in the supervised learning problem. This approximation has been considered in \cite{he2016improve}, including a discussion regarding the weights of the Gaussians. It should be noted that when the likelihood values are close, i.e., $P_{\hat{\theta_0}}(x^{n-1}) \approx P_{\hat{\theta}_1}(x^{n-1}) \approx \ldots \approx P_{\hat{\theta}_k}(x^{n-1})$, then a proper choice for the weights would be:

\begin{align*}
    w_i = \frac{\sqrt{I^{-1}_{\hat{\theta}_i}}}{\sum_{j=0}^k {\sqrt{I^{-1}_{\hat{\theta}_j}}}}.
\end{align*}

Now, note that under our assumptions, $I_{\hat{\theta}_0} \approx I_{\theta_0
}$ has $d-k$ eigenvalues that are very small, $\lambda_{k+1}, \dots \lambda_{d} \ll \frac{k}{4nR^2}$. Assuming that other local maxima of the likelihood do not have such a property, the weight of $w_0$ might be very close to $1$, which justifies an analysis that is based on the local properties around $\theta_0$.  

\section{Spectrum-Based Regret Bounds -- Details for Agnostic Case} \label{sec:auxiliary_complexity_measures_details_agnostic}
\subsection{Language Models -- Online Case}
We shall now turn our attention to the agnostic case, where the data-generating distribution $P(\cdot)$ might not be represented by any member of the hypothesis class $\Theta$. 
Let $P_{\theta_0}$ denote the model in the class closest to $P$ in KL divergence sense as defined in (\ref{equ:theta0_online}).
For the agnostic online case, the proof follows that of the realizable online case, but instead of using the Fisher information matrix we will use the Hessian of the log-loss at $\theta_0$, i.e.:
\begin{align}
    H^{o} & = - \mathbb{E}_{x^n \sim P(x^n)} \left[ \frac{\partial^2 \log P_{\theta_0}(x^n)}{\partial \theta^2} \mymid_{\theta_0} \right].
\end{align}

\begin{theorem}
\label{ausilary_online_agnostic_theorem}
Suppose $\Theta$ is contained within a $d$-dimensional ball of radius $R$, and the prior $w(\theta)$ is uniform over $\Theta$. Further assume that the eigenvalues of the Hessian matrix $H^{o}$, denoted $\lambda_1 \geq \lambda_2 \geq \cdots \geq \lambda_d$, satisfy
\[
\lambda_{k+1} \leq \frac{k}{R^2 \log(e)} \quad \text{for some } k.
\]
Then
\begin{align}
    \regret_a^o(Q,P) 
    &\leq \frac{1}{2n}\biggl[\frac{k}{\log(e)} 
        + k \log\!\left(\frac{\log(e)}{k}\right) 
        + \sum_{i=1}^k \log\!\bigl(R^2 \lambda_i\bigr)\biggr].
    \label{equ:online_regret_general_rad}
\end{align}
\end{theorem}

\begin{proof}
Write the local expansion of the log-loss around $\theta_0$ for the agnostic case. This is the natural generalization of the KL expansion in (\ref{equ:expansion}):
\begin{align}
\mathbb{E}_{x^n \sim P(x^n)} \left[ \log \left( \frac{P_{\theta_0}(x^n)}{P_{\theta}(x^n)} \right) \right] 
&= (\theta-\theta_0)^\top \frac{H^{o}}{2} (\theta-\theta_0) 
   + O\!\left(\|\theta-\theta_0\|^3\right).
\label{equ:expansion_agnostic}
\end{align}

Note that since $\theta_0$ attains the minimal log-loss, $\mathbb{E} \left[\nabla \log \left( P_{\theta_0}(x^n) \right) \right]^T = \vec{0}$ and thus the first term in Taylor's expansion, $\mathbb{E} \left[\nabla \log \left( P_{\theta_0}(x^n) \right) \right]^T (\theta - \theta_0)$ vanishes. The rest of the proof follows exactly the same steps as Theorem \ref{the:online_local_properties}, 
except that the Fisher Information Matrix is replaced with the Hessian.

Nevertheless, we repeat it here for the convenience of the reader. 
Let $d$ denote the dimensionality of $\Theta$, and let $\{\lambda_i\}_{i=1}^d$ 
be the ordered eigenvalues of $H^{o}$.
Let $d$ denote the dimensionality of $\Theta$, and let $\{\lambda_i\}_{i=1}^d$ be the ordered eigenvalues of $H^{o}$.  
From~\eqref{equ:expansion_agnostic}, ignoring higher-order terms, the set $\Theta_0$ is approximately an ellipsoid with principal axis lengths
\[
\left\{\frac{\sqrt{2n}\,\epsilon_n}{\sqrt{\lambda_i}}\right\}_{i=1}^{d},
\]
where $\epsilon_n$ captures the dependence on $n$. Hence the volume of $\Theta_0$, denoted $V(\Theta_0)$, is approximately
\begin{align}
    V(\Theta_0) 
        \;\cong\; \frac{(\sqrt{2n}\,\epsilon_n)^d}{\prod_{i=1}^{d} \sqrt{\lambda_i}} 
        \, V\!\left(S(1)\right),
\end{align}
where $S(1)$ is the unit $d$-dimensional ball and $V(\cdot)$ denotes volume.

Assume now that the prior is uniform over a $d$-dimensional volume contained in a ball of radius $R$ centered at $\theta_0$. Then
\[
w(\Theta_0) \;\geq\; \frac{V(\Theta_0 \cap S(R))}{V(S(R))}.
\]
If along some axis the ellipsoid radius exceeds $R$, the effective radius is capped at $R$. Thus the effective axes are
\[
\left\{\min\!\left\{\frac{\sqrt{2n}\,\epsilon_n}{\sqrt{\lambda_i}},\,R\right\}\right\}_{i=1}^d.
\]
Let $k$ be the largest index such that $\frac{\sqrt{2n}\,\epsilon_n}{\sqrt{\lambda_k}} \leq R$. We call $k=k(n)$ the {\em effective number of parameters}. It follows that
\begin{align}
\label{equ:ratio_agnostic}
w(\Theta_0) 
   &= \frac{V(\Theta_0 \cap S(R))}{V(S(R))} \\
   &\cong \frac{(\sqrt{2n}\,\epsilon_n)^k \, R^{d-k}}{\prod_{i=1}^k \sqrt{\lambda_i}} 
      \cdot \frac{V(S(1))}{R^d V(S(1))} \nonumber \\
   &= \frac{(\sqrt{2n}\,\epsilon_n)^k}{R^k \prod_{i=1}^k \sqrt{\lambda_i}}. \nonumber
\end{align}
Substituting~\eqref{equ:ratio_agnostic} into~\eqref{equ:online_general_bound_min_epsilon} gives
\begin{align}
\regret^o_a \;\leq\; 
    \min_{\epsilon_n} \Biggl[ 
        \epsilon_n^2 
        - \frac{k}{2n}\log(2n \epsilon_n^2) 
        + \frac{1}{2n}\sum_{i=1}^k \log(R^2 \lambda_i)
    \Biggr].
\end{align}
Differentiating with respect to $\epsilon_n^2$ shows the minimizer is
\[
\epsilon_n^2 \;=\; \frac{k}{2n\log(e)}.
\]
Substituting this back yields
\begin{align*}
\regret_a^o 
   &\leq \frac{1}{2n}\biggl[\frac{k}{\log(e)} 
        + k \log\!\left(\frac{\log(e)}{k}\right) 
        + \sum_{i=1}^k \log(R^2 \lambda_i)\biggr],
\end{align*}
as claimed.

\end{proof}

\subsection{Supervised Learning }
\label{sec:supervised_online_to_batch}

For supervised learning, rather than evaluating the bounds directly, we will use a clever method that was introduced, for example, in \cite{cesa2004generalization} which allows an ``online-to-batch'' conversion. To this end we employ a slightly different learner, namely:
\begin{align}
\label{equ:learner_online_to_batch_supervised}
    \tilde{Q}(y \mymid x ;\mathcal{S}) = \frac{1}{n+1}\sum^n_{t=0} Q(y \mymid x ; \mathcal{S}_t),
\end{align}
where $\mathcal{S}_t = \{(x_i, y_i)\}_{i=1}^{t} $, $\mathcal{S}_n = \mathcal{S}$ is the whole training set, $\mathcal{S}_0$ is an empty sequence, and $Q(\cdot)$ is a Bayesian probability assignment of the following form:
\begin{align}
    Q(y \mymid x; \mathcal{S}_t) = \frac{\int_{\theta} w(\theta) P_{\theta}( y \mymid x) \prod_{i=1}^tP_{\theta}(y_i \mymid x_i)}{\int_{\theta} w(\theta) \prod_{i=1}^tP_{\theta}(y_i \mymid x_i)}.
\end{align}
Note that this learner differs from our standard Bayesian mixture model in the sense that now we also employ an average ``over time.'' In principle, such a learner could also be implemented by modern systems by training many models, with an increasing number of data samples, and then averaging over all these models. Of course, this approach increases complexity, and intuitively, it should perform worse, but the approach leads to elegant proofs.
For the learner \eqref{equ:learner_online_to_batch_supervised}, we have the following result:
\begin{theorem}
\label{the:supervisedagnostic}
    Assume that the data-generating distribution is memoryless, i.e., $\forall{x^n, y^n}: P(y^n, x^n) = \prod_{t=1}^n P(y_t, x_t)$. Then the following holds:
    \begin{align}
        \regret^s_a(\tilde{Q},P; n)  \leq \frac{\regret_a^o(Q, P; n+1)}{n+1},
    \end{align}
    where $\regret^o(Q, P; n+1)$ is the regret for the online case with $n+1$ samples.  
\end{theorem}
\begin{proof}
First, note that since the same joint probability distribution $P(x,y)$ generates both the test samples $(x, y)$ and every pair in the training set $\mathcal{S} = \{(x_i,y_i)\}_{i=1}^n$, we get that :
\begin{align}
    \forall{t}: Q_{t+1} & \overset{\Delta}{=} \mathbb{E}_{\mathcal{S}, x, y}{\left[ \log \left( \int_{\theta} w(\theta) P_{\theta}( y \mymid x) \prod_{i=1}^tP_{\theta}(y_i \mymid x_i) \right) \right]} \notag \\
    & = \mathbb{E}_{\mathcal{S}, x, y}{\left[ \log \left( \int_{\theta} w(\theta) \prod_{i=1}^{t+1}P_{\theta}(y_i \mymid x_i) \right) \right]}.
\end{align}
In addition, since $ \forall \theta \in \Theta, n > 0: P_{\theta}(y^n \mymid x^n) = \prod_{t=1}^n P_{\theta}(y_t \mymid x_t)$, we have:
\begin{align}
    \mathbb{E} \left[ \sum_{t=1}^n \log{\left( P_{\theta}(y \mymid x) \right)} \right] = \mathbb{E} \left[ \log \left( P_{\theta}(y^n \mymid x^n) \right) \right].
\end{align}
Thus, we get:
\begin{align}
\begin{split}        
    \regret^s_a(\tilde{Q},P; n)  
    & \stackrel{(\ref{equ:regret_supervised_agnostic}) \& (\ref{equ:learner_online_to_batch_supervised})}{=} \mathbb{E}_{\mathcal{S} \sim P_{X} P, (x, y) \sim P_X P} \left[ \log \left( \frac{P_{\theta_0}(y \mymid x)}{\frac{1}{n +1}\sum^n_{t=0} Q(y \mymid x ; \mathcal{S}_t)} \right) \right]
    \\
    & \stackrel{\text{Jensen}}{\leq} 
    \frac{1}{n+1}\sum_{t=0}^n\mathbb{E}_{\mathcal{S} \sim P_{X} P, (x, y) \sim P_X P} \left[ \log \left( \frac{P_{\theta_0}(y \mymid x)}{Q(y \mymid x ; \mathcal{S}_t)} \right) \right]
    \\
    &=
    \frac{1}{n+1}\sum_{t=0}^n\mathbb{E}_{(x, y) \sim P_X P}\left[ \log \left(P_{\theta_0}(y \mymid x) \right) \right] - \frac{1}{n+1} \left( \sum_{t=1}^n \left( Q_{t+1} - Q_{t} \right) \right) 
    \\
    &= \frac{\mathbb{E}_{x^{n+1}, y^{n+1}} \left[ \log \left( {P_{\theta_0}}(y^{n+1} \mymid x^{n+1}) \right) -\log \left( Q(y^{n+} \mymid x^{n+1} \right) \right]}{n+1}
    \\
    & = \frac{\regret_a^o(Q, P; n+1)}{n+1}.
\end{split}
\end{align}
\end{proof}

Naturally, this result also holds for the agnostic batch-learning problem if both the hypothesis class and the underlying data-generating process are both memoryless, as in this case the batch learning problem boils down to supervised learning without data-features $x$.

\section{Comparison to Singular Learning Theory}
One branch in the literature that is especially relevant to our work is singular learning theory (SLT), see \cite{watanabe2010asymptotic}, \cite{watanabe2024recent} and references therein. This framework utilizes the Resolution Theorem by \cite{hironaka1964resolution} to characterize the asymptotic values of the regret for the online and batch learning problem for memoryless hypothesis classes, both in the realizable and the agnostic setting. Comparing our results in Appendix \ref{sec:regret_boundsdetails} to SLT, we give non-asymptotic upper bounds, which also hold for hypothesis classes and data-generating distributions with memory. It is interesting, however, to note that the interpretation of the real log canonical threshold as the $\lambda = \lim_{\epsilon \to 0} \frac{\log \left( w(\Theta_0) \right)}{\log(\epsilon^2)}$, can be easily incorporated into the online regret bound we have in Appendix \ref{sec:regret_boundsonline}, to get a result which is equivalent to Theorem \ref{the:online_local_properties}: Assume that indeed, $w(\Theta_0) = \alpha \epsilon^{2\lambda} + o(\epsilon)$ for some $\alpha > 0$. Plugging this into \eqref{equ:online_general_bound_min_epsilon}, we get:
\begin{align}
    \regret_a^o(Q, P) \leq \min_{\epsilon^2} \epsilon^2 - \frac{ \lambda\log(\alpha \cdot \epsilon)}{n}.
\end{align}
This bound has a minimum at $\epsilon^2 = \frac{\lambda}{n}$, which yields:

\begin{align}
    \regret_a^o(Q, P) \leq \frac{\lambda \log(n)}{n} +  \frac{\lambda (1+ 2 \log(\alpha) - 2 \log(\lambda)}{n} + o \left( \frac{1}{n} \right).
\end{align}


\section{The Role of the Architecture -- Details}
\subsection{FIM for Linear Regression}
\label{app:linear_fim}

Consider the linear regression model
\[
y = x^\top \theta + n,\quad x\sim P_X(\cdot), \quad n \sim \mathcal{N}(0,\sigma^2).
\]
The likelihood of $y$ given $x$ is
\[
P_\theta(y \mymid x) = \frac{1}{\sqrt{2\pi \sigma^2}}
\exp\!\left(-\frac{1}{2\sigma^2}(y - x^\top \theta)^2\right).
\]
Taking gradients with respect to $\theta$,
\[
\nabla_\theta \log P_\theta(y \mymid x) = \frac{1}{\sigma^2}(y - x^\top \theta)\, x.
\]
The FIM is
\begin{align}
\mathrm{FIM}
&= \mathbb{E}_{x \sim P_X(\cdot); \;\; n \sim \mathcal{N}} \big[ \nabla_\theta \log P_\theta(y\mymid x)\, \nabla_\theta \log P_\theta(y\mymid x)^\top \big] 
= \frac{1}{\sigma^2}\, \mathbb{E}_{x \sim P_X(\cdot)}[x x^\top].
\end{align}  
Thus, the FIM is simply the covariance matrix of the inputs and does not depend on $\theta$. For i.i.d.\ inputs the eigenvalues are constant, while for stationary processes they correspond to the Fourier spectrum. In either case there is no reason to expect degeneracy. A similar conclusion is reached in \citet{GluchUrbanke2023}.

\subsection{Spectral Properties of Products of Random Matrices}
\label{app:random_products}

Let $A_1,\dots,A_{\ell}$ be independent random matrices with i.i.d.\ complex entries of variance $1/\sqrt{w}$ (matrix width $w$). Define their product
\[
A = \prod_{i=1}^{\ell} A_i.
\]
As $w \to \infty$, the eigenvalue distribution of $A$ converges to a deterministic law \citep{GoetzeTikhomirov2011,BougerolLacroix2014}. Specifically, the eigenvalues lie in the unit disc with density
\[
g(x,y) = \frac{1}{\pi {\ell}\,(x^2+y^2)^{({\ell}-1)/{\ell}}}\,
\mathbf{1}_{\{x^2+y^2 \leq 1\}},
\]
where $(x,y)$ denote coordinates in the complex plane.  For the radial magnitude $r=\sqrt{x^2+y^2}$, the density is
\[
g(r) = \frac{2}{{\ell}\, r^{({\ell}-2)/{\ell}}} \,\mathbf{1}_{\{0 \le r \le 1\}}.
\]
Hence, the fraction of eigenvalues with $r \leq \delta$ is $\delta^{4/{\ell}}$, which tends to $1$ as ${\ell} \to \infty$.  
This shows that deeper factorizations bias the spectrum toward small eigenvalues, even though the overall function space (linear maps) remains unchanged. See also \citet{AroraEtAl2019} for connections to optimization and SGD convergence.

\subsection{FIM for Composition of Linear Maps}
\label{app:fim_composedlinearmaps}
Let us evaluate the FIM associated with the linear neural network above. Specifically, we define the dimensions:
\[
A_k \in \mathbb{R}^{d_{k-1} \times d_k},
\qquad
d_0 = \dim(y), \quad d_l = \dim(x).
\]
and let 
$Z_{>k} := A_{k+1} \cdots A_{{\ell}} x = B_k x\in \mathbb{R}^{d_k}
$ and 
$Z_{<k} := A_1 \cdots A_{k-1} \in \mathbb{R}^{d_0 \times d_k}.
$
The FIM can be written as a block matrix where the $i,j$ block corresponds to the parameters $A_i,A_j$ written in a vector notation, and 
\[
I_{ij}
\;=\;
\frac{1}{\sigma^2}\,
\big( B_i \,\Sigma_x\, B_j^\top \big)
\;\otimes\;
\big( Z_{<i}^\top Z_{<j} \big).
\]
Now, the eigenvalues of the FIM can be analyzed. Since the block terms are composed of the Kronecker product of nonnegative defined matrices, each composed of matrix multiplications, then as discussed above, the eigenvalues of each term will tend to degenerate and the Kronecker product further enhances that as its eigenvalues are the multiplication of all pairs of eigenvalues.

\subsection{FIM for General Layered Models}
\label{app:layered_fim}

Consider the composite model
\[
y = f_{\theta_1}(f_{\theta_2}(\cdots f_{\theta_\ell}(x)\cdots)) + n,
\quad x\sim P_X(\cdot), \quad n \sim \mathcal{N}(0,\sigma^2),
\]
Let $h_\ell = x$, $h_{k-1} = f_{\theta_k}(h_k)$, and $h_0 = f_\theta(x)$. Each layer has Jacobian
\[
J_k = \frac{\partial f_{\theta_k}(h_k)}{\partial h_k}.
\]
The Fisher Information Matrix decomposes into blocks indexed by parameters $\theta_i,\theta_j$:
\[
I_{ij} = \frac{1}{\sigma^2}\,\mathbb{E}_{x \sim P_X(\cdot)} \!\big[
\nabla_{\theta_i} f_\theta(x)\,
\nabla_{\theta_j} f_\theta(x)^\top
\big].
\]
By the chain rule,
\[
\nabla_{\theta_i} f_\theta(x)
= J_{1:i-1}\, \nabla_{\theta_i} f_{\theta_i}(h_i),
\qquad J_{1:i-1} := \prod_{k=1}^{i-1} J_k.
\]
Therefore,
\[
I_{ij} = \frac{1}{\sigma^2}\,
\mathbb{E}_{x \sim P_X(\cdot)}\!\left[
\Big(\prod_{k=1}^{i-1} J_k\Big)\,
\nabla_{\theta_i} f_{\theta_i}(h_i)\,
\big(\nabla_{\theta_j} f_{\theta_j}(h_j)\big)^\top\,
\Big(\prod_{k=1}^{j-1} J_k\Big)^\top
\right].
\]

As in the linear case, each block involves products of Jacobians, which tend to degenerate. This yields an FIM with many small eigenvalues, reflecting an implicit bias toward “simpler’’ effective functions.  

This formulation applies beyond the Gaussian case. For a general layered probabilistic model defined by conditional distributions
\[
P_{\theta_i}(h_i
\mymid h_{i-1}),\quad i=1,\dots,\ell,
\]
the FIM retains dependence on the layered structure via accumulated Jacobians, leading to the same qualitative bias. See \citet{KarakidaEtAl2021,Sun2025GeometricModeling} for further analysis in ReLU networks.

\section{The Zero Function has Small Complexity in a ReLU Network}
\label{app:zero_function_has_small_complexity}
In Section~\ref{sec:model_complexity} we discussed that distributions/functions that can be implemented with a small number of neural units have small complexity. Let us illustrate this point via a simple but concrete example.

\subsection{Setup}

Consider a two-layer ReLU network of width $m$:
\[
f_\theta(x) \;=\; a^\top \phi(Wx+b), \qquad 
\phi(z) = \max\{0,z\}\ \text{(applied coordinatewise)},
\]
with parameters $\theta=(W,b,a)$, where $W\in\mathbb{R}^{m\times d}$,
$b\in\mathbb{R}^m$, $a\in\mathbb{R}^m$. 
Inputs are restricted to a bounded domain $\mathcal{X}=\{x:\|x\|_2\le R\}$.
We place a uniform prior on a bounded parameter box
\[
W_{ij}\in[-W,W], \quad b_j\in[-B,B], \quad a_j\in[-A,A],
\]
independently across coordinates.

\subsection{Inactive Units and the Zero Function}

A hidden unit $j$ is said to be \emph{always inactive} on $\mathcal{X}$ if
\[
w_j^\top x + b_j \le 0 \qquad \text{for all } x\in\mathcal{X}.
\]
A sufficient condition is 
\[
b_j \;\le\; -R\|w_j\|_2.
\]
If every hidden unit is always inactive, then $\phi(Wx+b)=0$ for all $x\in\mathcal{X}$,
hence $f_\theta(x)\equiv 0$ \emph{regardless of $a$}. 
This describes a large-volume region of parameter space that collapses to the same
simple function.

The lemma below states that for a single unit to be inactive the prior probability is $p_{\mathrm{inact}}$. 
Because the units are independent under the prior, the probability that \emph{all}
$m$ hidden units are inactive is $p_{\mathrm{inact}}^m$. On this event, 
$f_\theta\equiv 0$ for all $a\in[-A,A]^m$.
\begin{lemma}[Prior mass of always inactive units]
Define 
\[
T = \min\!\left\{W,\; \tfrac{B}{R}\right\}, \qquad
V_d = \frac{\pi^{d/2}}{\Gamma(\tfrac{d}{2}+1)}
\ \text{(volume of the $d$-dimensional unit ball)}.
\]
Then the probability (under the uniform prior) that a single hidden unit is always inactive is
\[
p_{\mathrm{inact}}
= \frac{V_d}{(2W)^d(2B)}
\left(B\,T^d - \frac{dR}{d+1}\,T^{d+1}\right).
\]
\end{lemma}
\begin{proof}
Fix one hidden unit with parameters $(w,b)$ and assume the uniform prior on the
box $w\in[-W,W]^d$, $b\in[-B,B]$.
Let
\[
T \;=\; \min\!\left\{W,\;\frac{B}{R}\right\}.
\]
For any $w$ with $\|w\|_2=r\le T$ we have $-Rr \ge -B$, so the constraint
$b \le -Rr$ defines a nonempty interval within $[-B,B]$.

Given such a $w$ (with $\|w\|_2=r\le T$), the allowed $b$ values lie in
$[-B,\,-Rr]$. Hence the $b$-length is
\[
\mathrm{len}_b(r) \;=\; (-Rr) - (-B) \;=\; B - Rr.
\]
Let $S_d$ be the surface area of the unit sphere in $\mathbb{R}^d$ and
$V_d=\pi^{d/2}/\Gamma(\tfrac{d}{2}+1)$ the unit ball volume, with $S_d=d\,V_d$.
Using polar coordinates $w = r\,u$ with $r\in[0,T]$ and $u$ on the unit sphere,
the volume of the always-inactive set inside the box is
\[
\mu(\mathcal{S})
\;=\; \int_{\|w\|\le T} \mathrm{len}_b(\|w\|)\,dw
\;=\; \int_{\|w\|\le T} (B - R\|w\|)\,dw
\;=\; S_d \int_0^T (B - R r)\, r^{d-1}\,dr.
\]
Evaluating the integral,
\[
\mu(\mathcal{S})
\;=\; S_d\!\left( \frac{B\,T^d}{d} - \frac{R\,T^{d+1}}{d+1}\right)
\;=\; dV_d\!\left( \frac{B\,T^d}{d} - \frac{R\,T^{d+1}}{d+1}\right)
\;=\; V_d\!\left( B\,T^d - \frac{dR}{d+1}\,T^{d+1}\right).
\]
The prior box for $(w,b)$ has volume $(2W)^d\,(2B)$. Therefore the probability
that a single hidden unit is always inactive is
\[
p_{\mathrm{inact}}
\;=\; \frac{\mu(\mathcal{S})}{(2W)^d (2B)}
\;=\; \frac{V_d}{(2W)^d (2B)}
\left( B\,T^d - \frac{dR}{d+1}\,T^{d+1}\right).
\]
This expression is nonnegative since $T\le B/R$ ensures $B\,T^d \ge \tfrac{dR}{d+1}\,T^{d+1}$.

\end{proof}

\subsection{Implication for Complexity}

The prior mass of parameters yielding the zero function is therefore
\[
w(\Theta_{0}) \;\ge\; p_{\mathrm{inact}}^{\,m}.
\]
Consequently, the prior complexity, see (\ref{equ:pior_complexity}), satisfies
\[
\mathrm{Comp}_{\mathrm{prior}} \;\le\; -\,m\log p_{\mathrm{inact}}.
\]

\paragraph{Interpretation.}
This calculation shows explicitly that a very simple function (the zero function)
is supported by a large-volume region of parameter space: many different 
parameter choices collapse to the same output due to ReLU inactivity.
According to our framework, this yields low prior complexity and therefore
tight regret bounds.

The above example can be strengthened considerably. Assume that instead of a ReLU network with a single hidden layer, we consider a ReLU network with many, let us say $\ell$ hidden layers. By the same argument, the output of the first hidden layer is strictly zero for a substantial portion of the parameters of the first layer. For these choices of the parameters in the first layer, the parameters in the following $\ell-1$ layers can take on any values, all choices correspond to the zero function. This means that the prior probability of the zero function is at least $p_{\mathrm{inact}}^{\,m}$, independent of $\ell$.
\end{document}